\documentclass{article}
\usepackage[utf8]{inputenc}
\usepackage{graphicx}
\usepackage{subcaption}
\usepackage{booktabs}
\usepackage{amssymb}
\usepackage{float}
\usepackage{changepage}
\usepackage{multirow}
\usepackage{amsmath}
\usepackage{diagbox}
\usepackage{natbib}
\usepackage[toc,page]{appendix}
\usepackage{authblk}
\graphicspath{ {./img/} }

\title{RLBoost: Boosting Supervised Models using Deep Reinforcement Learning}

\author[1]{Eloy Anguiano Batanero}
\author[2]{Ángela Fernández Pascual}
\author[1]{Álvaro Barbero Jiménez}

\affil[1]{Instituto de Ingeniería del Conocimiento, C/ Francisco Tomás y Valiente 11, 28049, Madrid, Spain}
\affil[2]{Universidad Autónoma de Madrid \break C/ Francisco Tomás y Valiente 11, 28049, Madrid, Spain}

\begin{document}

%\begin{keyword}
%Data Quality \sep Data Valuation \sep Deep Learning \sep Reinforcement Learning \sep Supervised Learning \sep Multi-attention.

%\end{keyword}

\maketitle

\begin{abstract}
Data quality or data evaluation is sometimes a task as important as collecting a large volume of data when it comes to generating accurate artificial intelligence models. In fact, being able to evaluate the data can lead to a larger database that is better suited to a particular problem because we have the ability to filter out data obtained automatically of dubious quality. 
In this paper we present RLBoost, an algorithm that uses deep reinforcement learning strategies to evaluate a particular dataset and obtain a model capable of estimating the quality of any new data in order to improve the final predictive quality of a supervised learning model. This solution has the advantage that of being agnostic regarding the supervised model used and, through multi-attention strategies, takes into account the data in its context and not only individually.
The results of the article show that this model obtains better and more stable results than other state-of-the-art algorithms such as LOO, DataShapley or DVRL.
\end{abstract}

\section{Introduction}
Collecting large amounts of quality data is always desirable, if not essential for most successful use cases in the field of machine learning. However, sometimes this is not possible. Usually, we either have a large amount of data that we doubt whether it is of good quality, or we have little data that we know that have high quality but it becomes complicated to obtain more for the intended use case. That can be either because of the novelty of the use case to be treated or because of the difficulty of the acquisition of data. Additionally, there may be some cases where the cost of re-training a model needs to be optimized as some models are very expensive to train, so if you have a method of evaluating new data you can delegate that decision to when you obtain sufficient quality data. 

The evaluation of data quality turns out to be a problem that concerns the entire field of supervised learning and needs to be addressed because of the potential benefits it can bring. However, it is necessary to clarify that a data point is sometimes not good or bad individually, but depends on the origin or quality of the rest of the available data, so it is desirable to bring a more contextual approach to it. Indeed, we can formulate our problem as an optimization problem where, given $N$ samples, the objective is to select the best samples to the given task.

Fortunately, there is a field of machine learning that tries to optimize processes in order to maximize a metric (or reward) function. Reinforcement learning algorithms are essentially search algorithms that find a solution by trial and error. Therefore, using this kind of algorithms could be the key to solve the data valuation problem, knowing that it should be scalable to large amounts of data, as well as agnostic to the supervised learning model used.

This is why this article proposes a method of data valuation based on reinforcement learning that takes into account the information that a data sample can provide, both individually and contextually.

%Structure of the paper
This paper is organized as follows. In Section~\ref{sec:soa} we briefly describe the state of the art in data quality assessment for supervised models. We present the proposed method in Section~\ref{sec:proposal}, and the experimental results over several datasets are shown in Section~\ref{sec:exps}. The paper ends with some conclusions in Section~\ref{sec:concl}.

\section{State of the art}
\label{sec:soa}
There are previous approaches in the field of data quality assessment for 
supervised learning models. First, and a fairly common method, is the Leave One Out (LOO) approach. This technique attempts to isolate the effect of each data point on the training sample by simply removing it from the set and looking at the results it achieves. The worse the result achieved by removing one of the records, the greater the value of that data point is considered to be. While simple, the computational cost of this approach is in the order of $O(N^2)$ for a dataset with $N$ training samples. Nevertheless, this method has been reformulated in a computationally less inefficient way~\citep{Koh2017UnderstandingFunctions}.

On the other hand, and based on game theory, Shapley values can be defined~\citep{Lundberg2017APredictions}. Shapley values are a fairly widely used method in variable selection and explainability of supervised learning models. However, it has been seen in the literature that they can also be used for the particular use case of data quality assessment at ~\citet{Ghorbani2019DataLearning}. The original computational cost of Shapley values is $O(N!)$ where $N$ is the number of training samples, so it is necessary to perform a Monte Carlo sampling for cases where then number of items to be discriminated is high. In particular, in~\citet{Ghorbani2019DataLearning} they reduce this cost to $O(2^n)$ by reducing the problem to a cooperative game and using a truncated Monte Carlo (TMC) sampling strategy. 

Finally, there is also prior work on methods that use reinforcement learning techniques for the problem of data quality assessment such as Data Valuation using Reinforcement Learning (DVRL) \citep{Yoon2020DataLearning}. This method proved to be experimentally quite powerful, while featuring a computational complexity well below those discussed above (see \ref{appendix:time}). It approaches the objective as a multi-step trajectory problem, so it is forced to generate a reward that takes into account a moving window of scores within each episode versus the score of the actual step, being the episode a set of batches of the whole training dataset. This technique uses the REINFORCE algorithm~\citep{Williams:92} and treats each record individually to produce a probability value to be selected or not by a multinomial distribution.

%Also, it , while there are better alternatives in the current state of the art of reinforcement learning, such as PPO \cite{Schulman2017ProximalAlgorithms}. On the other hand, the evaluation of the data was done by , without taking into account that the same record may have different quality depending on the rest of the data to be evaluated. Finally, in the original DVRL 

Therefore, the current paper proposes a novel methodology based on a reinforcement learning approach improving over the one previously discussed, aimed at enhancing certain aspects of the aforementioned technique.

%This can cause the effect of each of the data samples on the improvement of the internal supervised learning model to be diluted along the agent's trajectories, making it more complicated to learn a good valuation strategy. 
\section{RLBoost method}
\label{sec:proposal}
The problem of selecting and sorting a collection of training data for its application in a supervised learning model can be addressed as an optimization problem, making the field of reinforcement learning particularly relevant. Specifically, the task involves selecting a subset of records (actions) from a given set (state) to optimize the model score (reward). To accomplish this, the paper delves into the domain of reinforcement learning, with a focus on policy gradient and Actor-Critic strategies-based agents.

Reinforcement learning\citep{Sutton1998} is the field that deals with the optimization of decision sequences along a trajectory. In this way, we would go on to define the following concepts. 

\begin{enumerate}
  \item A trajectory $T$ is composed of single steps $t \in T$.
  \item  During the trajectory $T$, decisions $a_t$ are made based on certain situations or states $s_t$ for which an immediate reward $r_t$ is obtained.
  \item The agent must adjust the parameters $\theta$ of a policy $\pi$ by which to decide what action to take in a given state, which will be denoted as $\pi_{\theta}(a_t|s_t)$.
  \item In the case of agents based on Actor-Critic strategies it will also be necessary to adjust the parameters $\phi$ of a complementary estimator of the value function $V$ depending only on the current state. This estimator will be in charge of predicting the cumulative reward from the current state to the end of the trajectory, and such estimation will be noted in the form $V_{\phi}(s_t)$.
\end{enumerate}

\subsection{Proximal Policy Optimization}
The reinforcement learning algorithm used in this study is Proximal Policy Optimization (PPO). 
At ~\citet{Yoon2020DataLearning} they use a regular Policy Gradient method, but PPO~\citep{Schulman2017ProximalAlgorithms} usually outperforms these techniques by several reasons. First, we have the use of advantages $A_{\phi}(s_t, a_t, \pi) = \delta_t + \gamma \delta_{t+1} + \gamma^{2} \delta_{t+2} +... + \gamma^{T-2} \delta_{t+T-1}$ where $\delta_{t} = r_t + \gamma V_{\phi}(s_{t+1})-V_{\phi}(s_t)$ and $\gamma$ is defined as a discount parameter over the steps in a trajectory. Those advantages are used to check whether some actions can obtain higher/lower rewards than expected, and therefore they are reinforced/disincentivized. This leads us to a generalization of the advantages calculation called Generalized Advantages Estimation (GAE) defined as $A^{GAE(\gamma, \lambda)}_{t} = \sum\limits^{\infty}_{t=0} (\lambda \gamma)^t \delta_{t+1}$. Here, the use of an exponential weight discount $\lambda$ is needed to control the bias variance trade-off.

On the other hand, PPO incorporates an intrinsic reward mechanism via an entropy term to promote exploration. Finally, the clipping component of this algorithm's policy loss acts as a regularizer of the model, ensuring that the policy does not suffer from excessive changes between each iteration and thereby enabling smoother improvements to be made.

The original PPO algorithm has this loss function definition
\begin{equation} \label{eq:loss_ppo}
   L(\theta)=L^{clip}(\theta) -c_1 L^{VF}(\phi)+ c_2 S(\theta),
\end{equation}
with the following entropy bonus loss to encourage exploration
\begin{equation*} \label{eq:entropy_old}
   S(\theta) = \mathbb{E}_t\left[-\pi_\theta (a_t, s_t) \log(\pi_\theta(a_t, s_t))\right],
\end{equation*}
the value function (or critic) loss defined as
\begin{equation} \label{eq:loss_vf_old}
    L^{VF}(\phi) =  \mathbb{E}_t \left \| r_t + \gamma V_{\widehat{\phi}}(s_{t+1}, \pi)) -V_{\phi}(s_{t}, \pi)) \right \|^{2}_2,
\end{equation}
with an old copy of the value function estimator with parameters $\widehat{\theta}$, and the policy (or actor) loss as
\begin{equation} \label{eq:loss_pf_old}
    \begin{gathered}
        t_1 = A_{\phi}(s_t, a_t, \pi) R_t(\theta), \\
        t_2 = A_{\phi}(s_t, a_t, \pi) clip(R_t(\theta), 1-\epsilon, 1+\epsilon),\\
        L^{clip}(\theta) =  \mathbb{E}_t[\min(t_1, t_2)],
    \end{gathered}
\end{equation}
where $R_t = \pi_{\theta}(a_t, s_t)/\pi_{\widehat{\theta}}(a_t, s_t)$. Here, we can see that the clipping method is used as a kind of regularizer to avoid aggressive changes on the policy function as previously mentioned.

\subsection{PPO as a bandit agent}
After reviewing the existing literature on data evaluation using reinforcement learning, it became apparent that a simplified approach could potentially establish a starting point for improvement. As outlined in earlier sections, the problem was formulated as a one-step optimization problem in which a reinforcement learning algorithm selects a subset of data from a training dataset to improve a chosen metric. The size of the data batch for the agent's states ($N$) must be large enough to be representative of the original data, but not so large as to excessively increase the complexity of the problem.

As part of this simplification process, a reward mechanism was defined based on calculating the difference between the validation score obtained by a supervised estimator trained with the full data batch and the validation score of the same estimator with the subset of data selected by the reinforcement learning agent.

Having specified the use case at hand, one could argue that a bandit method could better fit in this case, since the only actions available in the environment are the selection of data points in the batch being evaluated, for which an immediate reward is obtained. 
But in fact, to adapt PPO to a problem formulated in this way, we only need to check that it is mathematically possible and thus take advantage of the improvements of the algorithm mentioned above. This adaptation involves making two assumptions: $\gamma = 0$ and $t = 0$ as there is no trajectory.

These assumptions lead us to rewrite the previous formulation. First, the advantages calculated in the GAE are simplified in the form:
\begin{equation*} \label{eq:gae_new}
    A_{\phi}(s, a, \pi) = \delta = r -V_{\phi}(s, \pi).
\end{equation*}
Therefore, the new expression of the policy loss is as follows
\begin{equation} \label{eq:loss_pf_new}
    \begin{gathered}
        t_1 = \delta R(\theta), \\
        t_2 = \delta clip(R(\theta), 1-\epsilon, 1+\epsilon),\\
        L^{clip}(\theta) =  \min(t_1, t_2),
    \end{gathered}    
\end{equation}

and the loss function of the value function (the critic part) is given by:
\begin{equation} \label{eq:loss_vf_new}
    L^{VF}(\phi) =  \left \| r - V_{\phi}(s, \pi) \right \|^{2}_2.
\end{equation}

It is worth noting that, as we see in the differences between equation \eqref{eq:loss_vf_old} and equation \eqref{eq:loss_vf_new}, what was previously a temporal difference error between the actual value functions estimates and the estimation of the value function of the next state has become only the estimate of the actual state. This tells us that the value function of this agent indicates whether a higher or lower difference at model score is expected exclusively for the actual data.

Finally, it should also be noted that the entropy bonus, which is the value that allows the agent to be able to explore, would look like this

\begin{equation*} \label{eq:entropy_new}
    S(\theta) = -\pi_\theta (a, s) \log(\pi_\theta(a, s)).
\end{equation*}

After having tested the mathematical feasibility of making the proposed change, it has been proven that not only the change is feasible but it also provides us with some interesting advantages for the specific use case, such as the estimation of the profits of filtering non-quality data at each step through the agent's critic estimation or the clipping of the actor (equation \eqref{eq:loss_pf_new}) as a method of regularization of the agent. 

\subsection{Model architecture}
\label{subsec:arq}
\begin{figure}[H]
    \begin{adjustwidth}{-2cm}{-2cm}
        \centering
        \includegraphics[width=\textwidth]{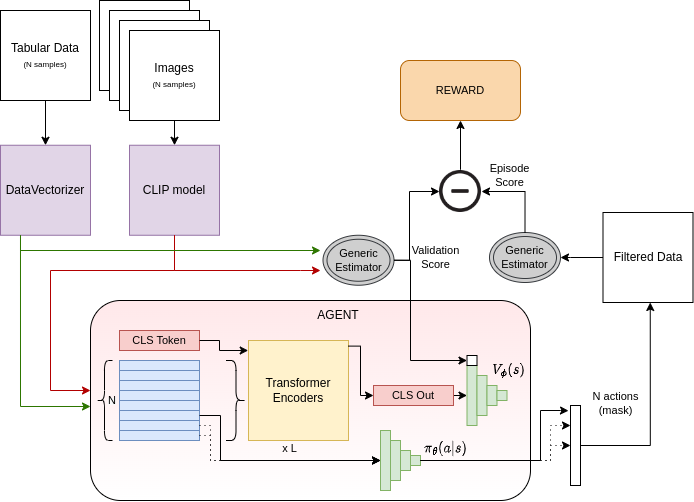}
    \end{adjustwidth}
    \caption{Schematic diagram of the operation of the proposed RLBoost model. It shows the application cases for both tabular data and image data, where a non-trainable data vectorizer will be needed for the latter. In this diagram it can be also seen that the reward calculation is the difference between the score of the episode using the proposed filtering versus the score of the model without filtering, both against the validation set.}
    \label{fig:RLBoost}
\end{figure}
For a complete understanding of the architecture of the model used by this particular algorithm and presented in Figure~\ref{fig:RLBoost}, it is necessary to clarify certain aspects beforehand.

First, depending on the use case for which the strategy presented in this paper is to be used, some kind of vectorization of each of the data samples to be evaluated is needed. In the case of tabular data this step is straightforward since each of the records is a vector in itself and the vectorizing part can be simply a series of fully connected layers or just one that functions as a projection to the latent state of the data. However, in the case of images, some kind of architecture would be needed to vectorize each of these samples, for example, a Contrastive Languaje-Image Pretrained model (CLIP)~\citep{Radford2021LearningSupervision}.

Once all the data has been vectorized, each of these vectors is used as input to a sequence of $L$ Transformer Encoders ~\citep{Vaswani2017AttentionNeed}. An extra parameterizable vector, known as \textit{CLS Token}, is also added to this process. The strategy of including a learnable vector called \textit{CLS} is frequently employed in document classification within the field of natural language processing (\textit{NLP})~\citep{Devlin2018BERT:Understanding}. The output of the $L$ Transformer Encoders corresponding to the \textit{CLS Token} is expected to capture the contextual information of the entire batch of vectors, allowing it to estimate the value function \eqref{eq:loss_vf_new}. In this case, the value function pertains to the complete batch of proposed data.
\section{Experiments}
\label{sec:exps}
As depicted in Figure \ref{fig:RLBoost}, the proposed method is versatile and applicable to various use cases, as long as the input data can be vectorized to serve as input to the Transformer Encoder. Hence, this approach will be evaluated on both tabular and image data, where the image data will be vectorized beforehand.

In addition, in order to test the training data filtering behavior that the proposed method can perform, a noise rate has been introduced in each of the train datasets to be tested. This error rate consisted of a target class shift to the opposite class of a random percentage of the training data at tabular data and a complete circular shift at image data (MNIST dataset), but neither at validation nor test datasets.

Since we are introducing noise into datasets that are expected to be sufficiently clean it has been decided to evaluate each of the proposed methods in two different ways. In all cases, we perform a test of the accuracy score of the supervised learning model using the filtering proposed by RLBoost against the same supervised learning model trained with all the available records, which we refer to as the ''baseline''.Additionally, as in some cases we are intentionally introducing noise ($15\%$ and $30\%$) we have also decided to evaluate the ability of the selection model to detect noisy records. Since in this case we also want to measure the ability of the model to detect data in which noise has been introduced, the evaluation of the data can be seen as a classification problem between noiseless and noisy records, such that the positive class is noiseless record and the negative class is noisy record. In this way, Receiver Operating Characteristic (ROC) curves can be calculated to measure the effectiveness of the method.

The methods compared in this study will be those presented in Section \ref{sec:soa}. The LOO, Data Shapley (SHAP), DVRL and RLBoost methods will be run for each of the use cases presented. It should be noted that the version of Data Shapley used needs to be the one that improves the computational cost of the algorithm from $O(N!)$ to $O(2^n)$ in order to obtain results for large dataset sizes.

It is necessary to emphasize that, since the LOO and SHAP methods are not agents intended for the task of data selection, a subsequent process has been performed after obtaining the data values from these methods. This process involves a sweep with different thresholds for the values obtained so that, once the best cut-off threshold is selected against the validation set, this threshold is used to train the final model.

This sweep has been performed in order to compare the rest of the methods against the best possible execution of these methods. However, it generates a couple of side-effects that must be taken into account:
\begin{itemize}
    \item  This sweep is very costly to perform in the face of new data arrivals as it forces several re-trainings , so it is not desirable to perform.
    \item The score results of the models using sweeping are measured in a way that generates a lower bound for the final score with the value of the baseline model. This is because a threshold that doesn't filter any data will always result in the baseline score. Therefore, the worst-case scenario for this method can only be the same as the baseline accuracy. In other words, the baseline score will be achieved even if no method had been applied.
\end{itemize}

\subsection{Tabular data}
Since each record in tabular datasets corresponds to a vector, the projection operation for this kind of problems is fairly straightforward. Specifically, it has been decided to use the Adult dataset and A5A, W5A and CIFAR10 datasets (from the LibSVM repository~\citet{Chang2011LIBSVM:Machines}) to test the performance with this kind of data. 

It should be noted that in the case of the CIFAR10 dataset, which is not a binary classification problem, the problem was previously binarized by simply segregating class \#1 from the rest. In all cases, the noise rates proposed for each of the tests are $ 0\% $, $ 15\%$  and $ 30\% $.

\begin{table}[H]
    \centering
    \begin{tabular}{c|ccc|cc}
    \toprule
    \textbf{Dataset} & \textbf{Train} & \textbf{Validation} & \textbf{Test} & \textbf{\# Features} & \textbf{\# Classes} \\ \midrule
    Adult & 1k & 400 & 47,442 & 123 & 2 \\ 
    A5A & 5k & 1k & 10k & 123 & 2 \\ 
    W5A & 5k & 1k & 10k & 300 & 2 \\ 
    CIFAR10 & 4k & 1k & 10k & 3,072 & 10 \textrightarrow 2\\ \bottomrule
    \end{tabular}
    \caption{Details of the records, features and classes of the tabular datasets to be used in the experimentation.}
    \label{table:tabular_datasets}
\end{table}

As can be seen in Table \ref{table:tabular_datasets}, in all cases priority has been given to having as large a dataset as possible under test to ensure that the evaluation of the algorithm performance is as realistic as possible. On the other hand, it should be noticed that in the case of the Adult dataset, the same training, validation and test cuts and the same preprocessing as in the original DVRL paper~\citep{Yoon2020DataLearning} have been performed.

The parameters used in each of the runs are common to all experiments. In all cases, the internal generic estimator was a logistic regression (to speed up the execution of the tests), the batch sizes of records proposed to the agent were of size 200 samples, the agent has 4 stacked Transformer Encoders, trained during 1e5 steps and the agent training batch being 64 in all proposed datasets.

% Adult Scores
\begin{table}[H]
    \centering
    \footnotesize

\begin{tabular}{@{}l||lll@{}|}
                                        & \textbf{0\% noise}      & \textbf{15\% noise}     & \textbf{30\% noise}     \\ \midrule
\textbf{Baseline}                       & $\mathbf{0.832}$  & $\mathbf{0.814}$  & $0.776$           \\
\textbf{LOO}                            & $0.832(\pm0.0)$   & $0.814(\pm0.0)$   & $0.776(\pm0.0)$   \\
\textbf{SHAP}                           & $0.832(\pm0.0)$   & $0.814(\pm0.0)$   & $0.776(\pm0.0)$   \\
\textbf{DVRL}                           & $0.832(\pm0.000)$ & $0.793(\pm0.029)$ & $0.768(\pm0.007)$ \\ \midrule
\textbf{RLB (1e-1)}                     & $0.807(\pm0.006)$ & $0.799(\pm0.008)$ & $0.813(\pm0.004)$ \\
\textbf{RLB (1e-2)}                     & $0.814(\pm0.005)$ & $0.811(\pm0.006)$ & $0.812(\pm0.007)$ \\
\textbf{RLB (1e-3)}                     & $0.813(\pm0.001)$ & $0.816(\pm0.004)$ & $\mathbf{0.814(\pm0.001)}$ \\
\textbf{RLB (1e-4)}                     & $0.814(\pm0.002)$ & $0.812(\pm0.005)$ & $0.814(\pm0.007)$ \\\bottomrule
\end{tabular}
\label{tab:scores_adult}

    \caption{Table of scores against test data using the final filtering proposed by each of the models at Adult dataset at 0\%, 15\% and 30\% noise ratio.}
    \label{tab:adult_scores}
\end{table}

% A5A Scores
\begin{table}[H]
    \footnotesize
    \centering

\begin{tabular}{@{}l||lll@{}|}
                                        & \textbf{0\% noise}      & \textbf{15\% noise}     & \textbf{30\% noise}              \\ \midrule
\textbf{Baseline}                       & $\mathbf{0.844}$  & $\mathbf{0.838}$  & $0.807$                    \\
\textbf{LOO}                            & $0.844(\pm0.0)$   & $0.838(\pm0.0)$   & $0.807(\pm0.0)$            \\
\textbf{SHAP}                           & -----             & -----             & -----                      \\
\textbf{DVRL}                           & $0.828(\pm0.002)$ & $0.829(\pm0.004)$ & $0.809(\pm0.016)$          \\ \midrule
\textbf{RLB (1e-1)}                     & $0.834(\pm0.004)$ & $0.831(\pm0.004)$ & $0.826(\pm0.006)$          \\
\textbf{RLB (1e-2)}                     & $0.833(\pm0.002)$ & $0.830(\pm0.004)$ & $\mathbf{0.831(\pm0.003)}$ \\
\textbf{RLB (1e-3)}                     & $0.827(\pm0.003)$ & $0.826(\pm0.002)$ & $0.823(\pm0.006)$        \\
\textbf{RLB (1e-4)}                     & $0.826(\pm0.002)$ & $0.825(\pm0.004)$ & $0.826(\pm0.001)$          \\ \bottomrule

\end{tabular}
\label{tab:scores_a5a}

    \caption{Table of scores against test data using the final filtering proposed by each of the models at A5A dataset at 0\%, 15\% and 30\% noise ratio.}
    \label{tab:a5a_scores}
\end{table}

% W5A Scores
\begin{table}[H]
    \footnotesize
    \centering
    
\begin{tabular}{@{}l||lll@{}|}
                                        & \textbf{0\% noise}      & \textbf{15\% noise}     & \textbf{30\% noise}               \\ \midrule
\textbf{Baseline}                       & $\mathbf{0.982}$  & $\mathbf{0.977}$  & $0.946$                     \\
\textbf{LOO}                            & $0.982(\pm0.0)$   & $0.977(\pm0.0)$   & $0.946(\pm0.0)$             \\
\textbf{SHAP}                           & -----             & -----             & -----                       \\
\textbf{DVRL}                           & $0.979(\pm0.004)$ & $0.970(\pm0.004)$ & $0.899(\pm0.096)$           \\ \midrule
\textbf{RLB (1e-1)}                     & $0.979(\pm0.001)$ & $0.970(\pm0.007)$ & $0.959(\pm0.005)$           \\
\textbf{RLB (1e-2)}                     & $0.974(\pm0.006)$ & $0.972(\pm0.005)$ & $\mathbf{0.966(\pm0.005)}$  \\
\textbf{RLB (1e-3)}                     & $0.980(\pm0.001)$ & $0.969(\pm0.004)$ & $0.961(\pm0.002)$           \\
\textbf{RLB (1e-4)}                     & $0.981(\pm0.001)$ & $0.970(\pm0.005)$ & $0.957(\pm0.005)$           \\ \bottomrule
\end{tabular}
\label{tab:scores_w5a}

    \caption{Table of scores against test data using the final filtering proposed by each of the models at W5A dataset at 0\%, 15\% and 30\% noise ratio.}
    \label{tab:w5a_scores}
\end{table}

% CIFAR10 Scores
\begin{table}[H]
    \footnotesize
    \centering
    
\begin{tabular}{@{}l||lll@{}|}
                                        & \textbf{0\% noise}      & \textbf{15\% noise}     & \textbf{30\% noise}                                 \\ \midrule
\textbf{Baseline}                       & $0.880$           & $0.781$           & $0.683$                                       \\
\textbf{LOO}                            & $0.890(\pm0.0)$   & $0.759(\pm0.0)$   & $0.648(\pm0.0)$                               \\
\textbf{SHAP}                           & -----             & -----             & -----                                         \\
\textbf{DVRL}                           & $0.866(\pm0.031)$ & $0.748(\pm0.017)$ & $0.648(\pm0.010)$                             \\ \midrule
\textbf{RLB (1e-1)}                     & $0.896(\pm0.003)$ & $0.855(\pm0.007)$ & $0.797(\pm0.007)$                             \\
\textbf{RLB (1e-2)}                     & $0.902(\pm0.006)$ & $0.869(\pm0.016)$ & $0.860(\pm0.007)$                             \\
\textbf{RLB (1e-3)}                     & $\mathbf{0.902(\pm0.002)}$ & $\mathbf{0.882(\pm0.014)}$ & $\mathbf{0.879(\pm0.011)}$  \\
\textbf{RLB (1e-4)}                     & $0.898(\pm0.004)$ & $0.873(\pm0.014)$ & $0.866(\pm0.007)$                             \\ \bottomrule
\end{tabular}
\label{tab:scores_cifar10}

    \caption{Table of scores against test data using the final filtering proposed by each of the models at binarized CIFAR10 dataset at 0\%, 15\% and 30\% noise ratio.}
    \label{tab:cifar_scores}
\end{table}

The tables \ref{tab:adult_scores}, \ref{tab:a5a_scores}, \ref{tab:w5a_scores} and \ref{tab:cifar_scores} show the results of each of the proposed methods for the indicated datasets. On the one hand we have the results of previous filtering methods such as LOO, SHAP and DVRL together with the model without any filtering (Baseline) and on the other hand we have the results of our RLBoost method (RLB) with different entropy bonus values (the hyperparameter that quantifies the exploration in the PPO algorithm).

To assess the stability of the aforementioned methods, 5 executions were performed for each method in order to calculate the standard deviation of the metric being evaluated. This allowed us to examine the variability and consistency of the results obtained. It is important to note that in the case of DVRL, several re-executions were required due to the need to make a final decision on whether or not to choose a datum, as it was not necessary to use the sweep of methods that did choose an action. However, in some cases, this decision was incompatible with the training of a final estimator, requiring a rerun. To obtain the 5 valid runs for all the cases proposed, 38 backup runs were necessary in addition to the 75 runs required (also considering the results of Table \ref{table:img_scores}).

Several things must be emphasized in this table. On the one hand it can be seen that the execution of the SHAP method on datasets where the training data size has 5k examples could not be performed since the computational efficiency of the algorithm does not make it feasible (see appendix \ref{appendix:time}). On the other hand, and as it is obvious, the baseline models worsen their performance as noise is introduced to the training dataset, which corroborates that the noise generation is done correctly.

Additionally, it can be also observed how the RLBoost-based methods outperform in cases where the noise is high and still work pretty well in cases where there is not noise, so it seems that it is a fairly robust method as far as improving the final metric of the model  (accuracy in our cases) is concerned.

% Scores
\begin{figure}[H]
    \begin{adjustwidth}{-1.5cm}{-1.5cm}
        \centering
        \includegraphics[width=\textwidth]{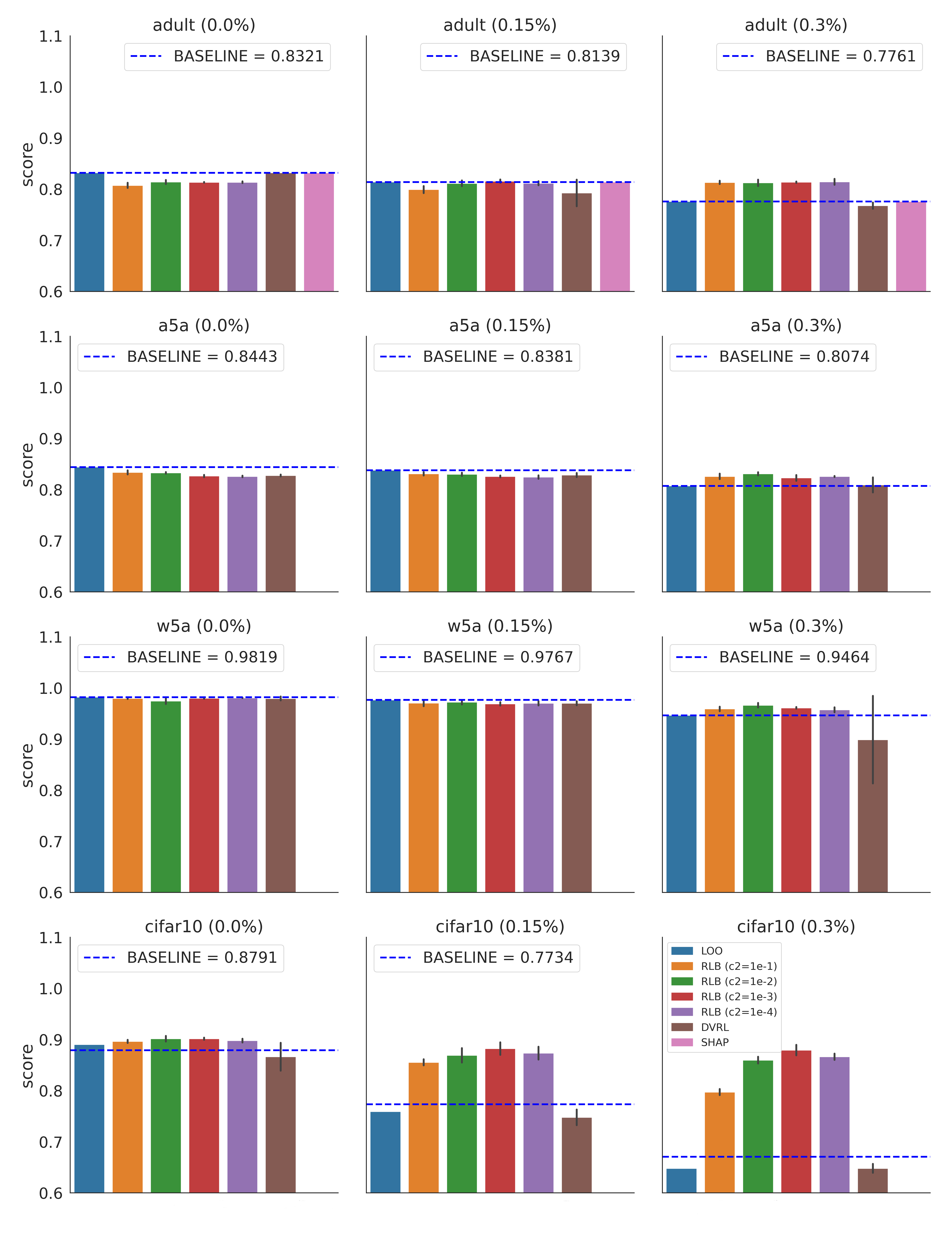}
        \caption{Final Scores for the agents trained over tabular datasets}
        \label{fig:tab_scores}
    \end{adjustwidth}
\end{figure}

As we can see, Figure \ref{fig:tab_scores} shows the performance against the test data for each of the models (0\% error, 15\% and 30\%) at the end of the final selection at every tabular dataset case tested. 

However, it can also be seen that the higher the error introduced, the more the generic estimator can take advantage of the data thanks to the selection made by our agent, which was the objective we were looking for.

% ROCs
\begin{figure}[H]
    \begin{adjustwidth}{-2cm}{-2cm}
        \centering
        \begin{subfigure}{0.65\textwidth}
            \centering
            \includegraphics[width=\textwidth]{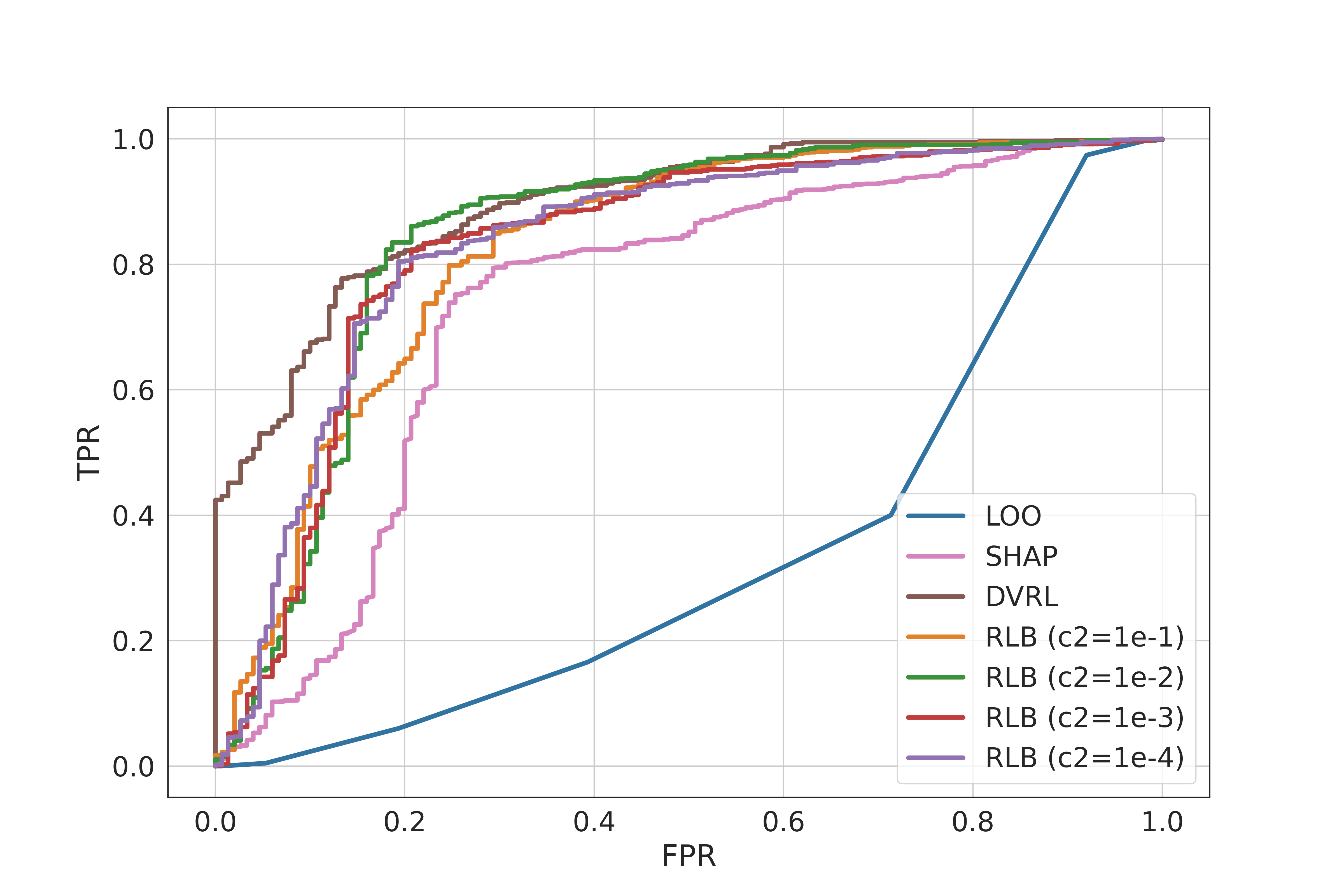}
            \caption{15\% error rate, best run in terms of AUC for each method.}
            \label{fig:roc_adult_15}
        \end{subfigure}
        \hfill
        \begin{subfigure}{0.65\textwidth}
            \centering
            \includegraphics[width=\textwidth]{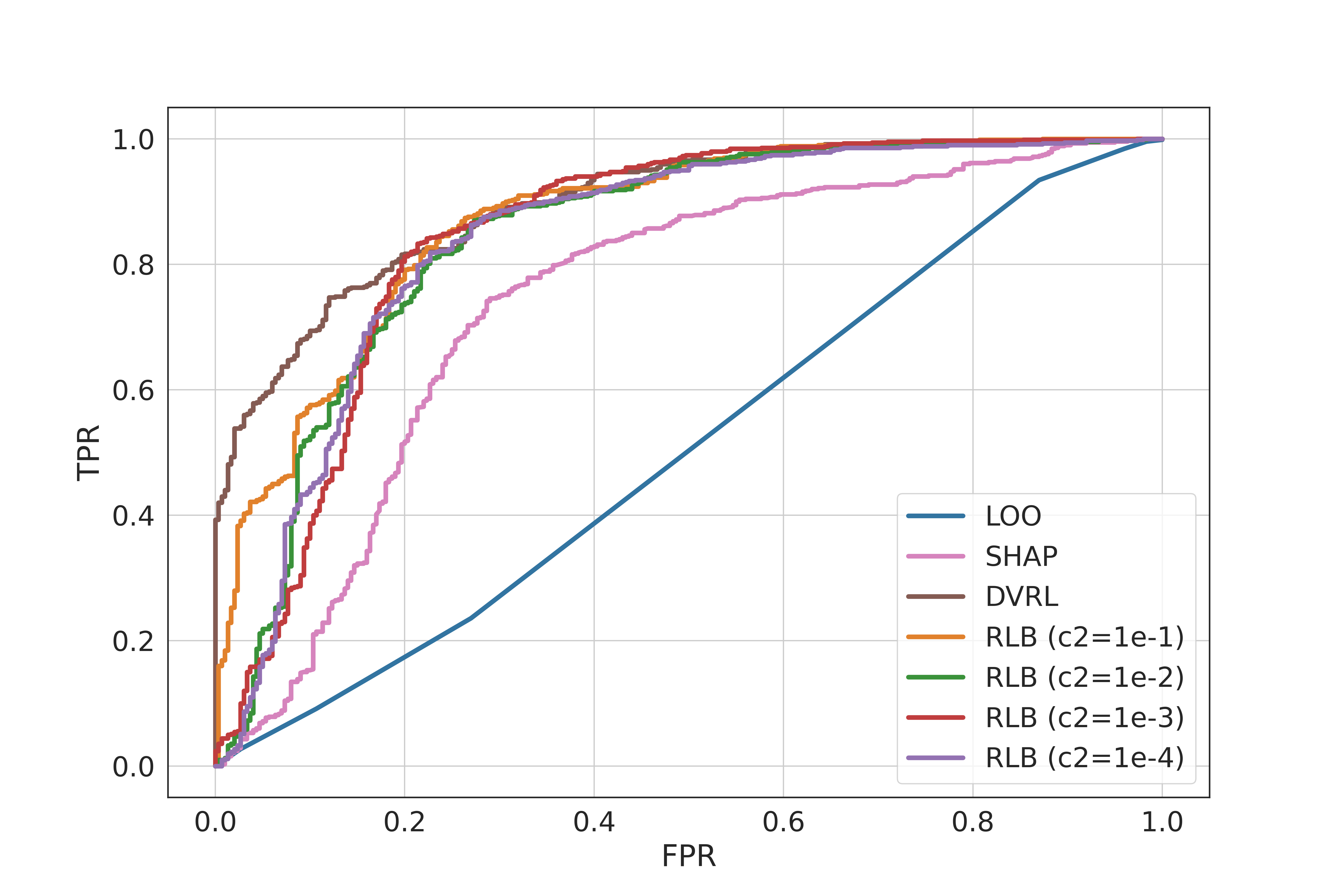}
            \caption{30\% error rate, best run in terms of AUC for each method.}
            \label{fig:roc_adult_3}
        \end{subfigure}
        \begin{subfigure}{0.65\textwidth}
            \centering
            \includegraphics[width=\textwidth]{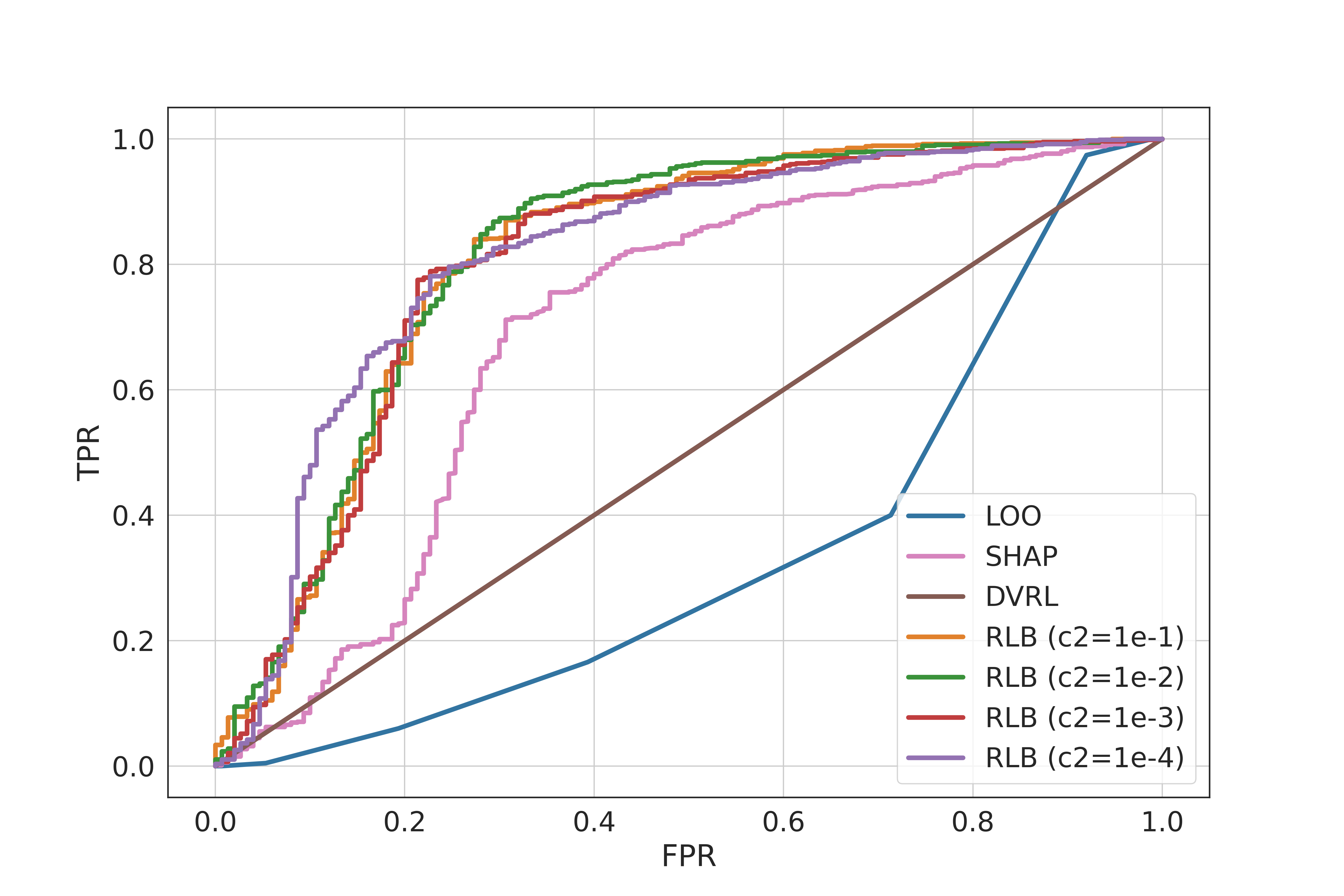}
            \caption{15\% error rate, worst run in terms of AUC for each method.}
            \label{fig:roc_adult_15_wrong}
        \end{subfigure}
        \hfill
        \begin{subfigure}{0.65\textwidth}
            \centering
            \includegraphics[width=\textwidth]{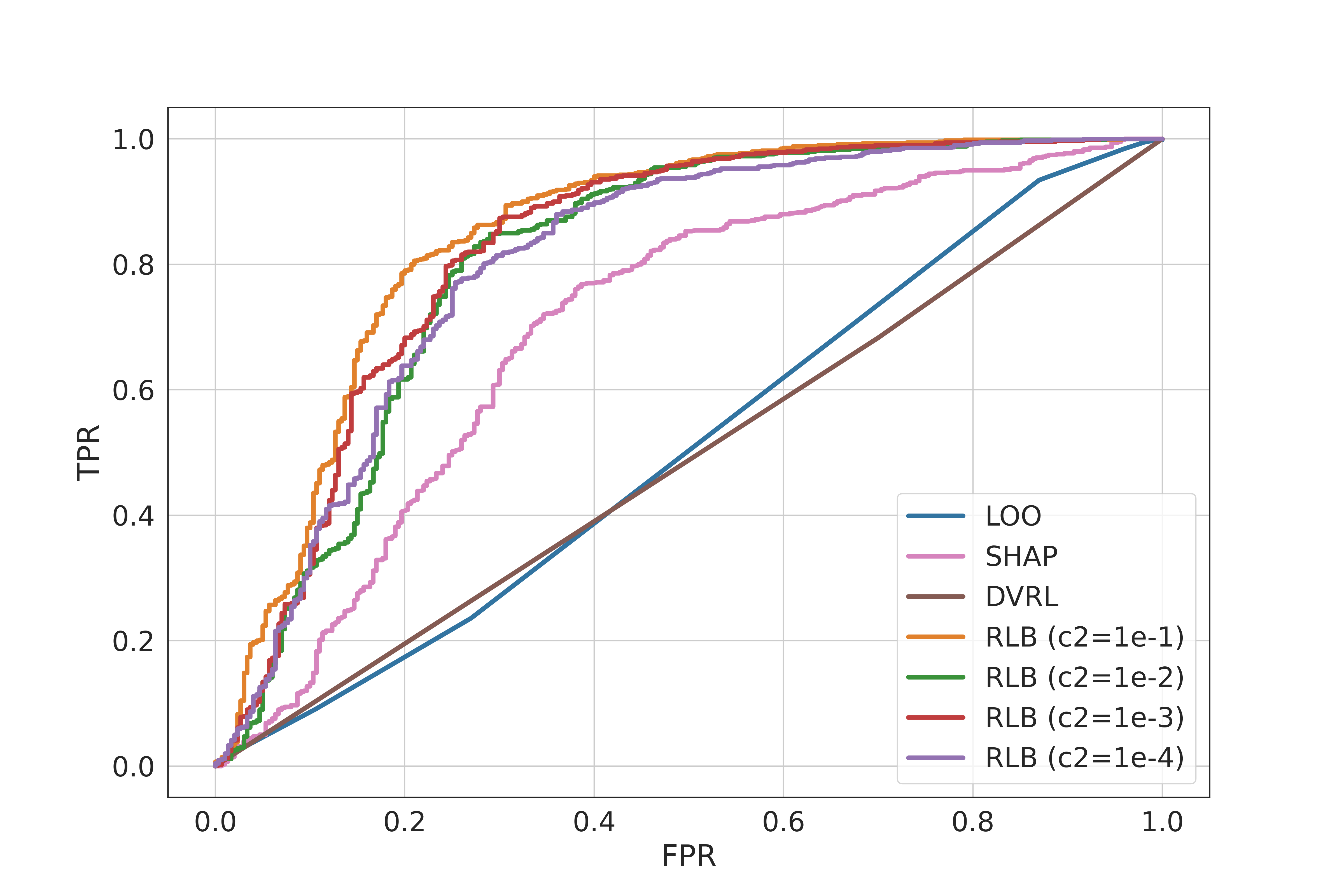}
            \caption{30\% error rate, worst run in terms of AUC for each method.}
            \label{fig:roc_adult_3_wrong}
        \end{subfigure} 
    \caption{ROC curves to measure the ability to detect noisy data in the Adult dataset, where the positive class is the data without noise and the negative class is the data with noise.}
    \label{fig:rocs_adult}
    \end{adjustwidth}
\end{figure}

On the other hand, in Figure \ref{fig:rocs_adult} and tables \ref{table:tab_aucs_adult}, \ref{table:tab_aucs_a5a}, \ref{table:tab_aucs_W5a} and \ref{table:tab_aucs_cifar} , we can see the ROC curves on whether the model has identified that a particular record is a data sample without noise (positive value), or it is a value with noise (negative value) based on the outputs of the agent before being binarized to be selected or not selected.

These curves show us that the capacity of the proposed agents are similar to the best of those available in the state of the art, but without the computational limitation of the execution of Data Shapley~\citep{Ghorbani2019DataLearning}, as well as the improved stability of our method over DVRL. 

\begin{figure}[H]
    \begin{adjustwidth}{-2cm}{-2cm}
        \centering
        \begin{subfigure}{0.65\textwidth}
            \centering
            \includegraphics[width=\textwidth]{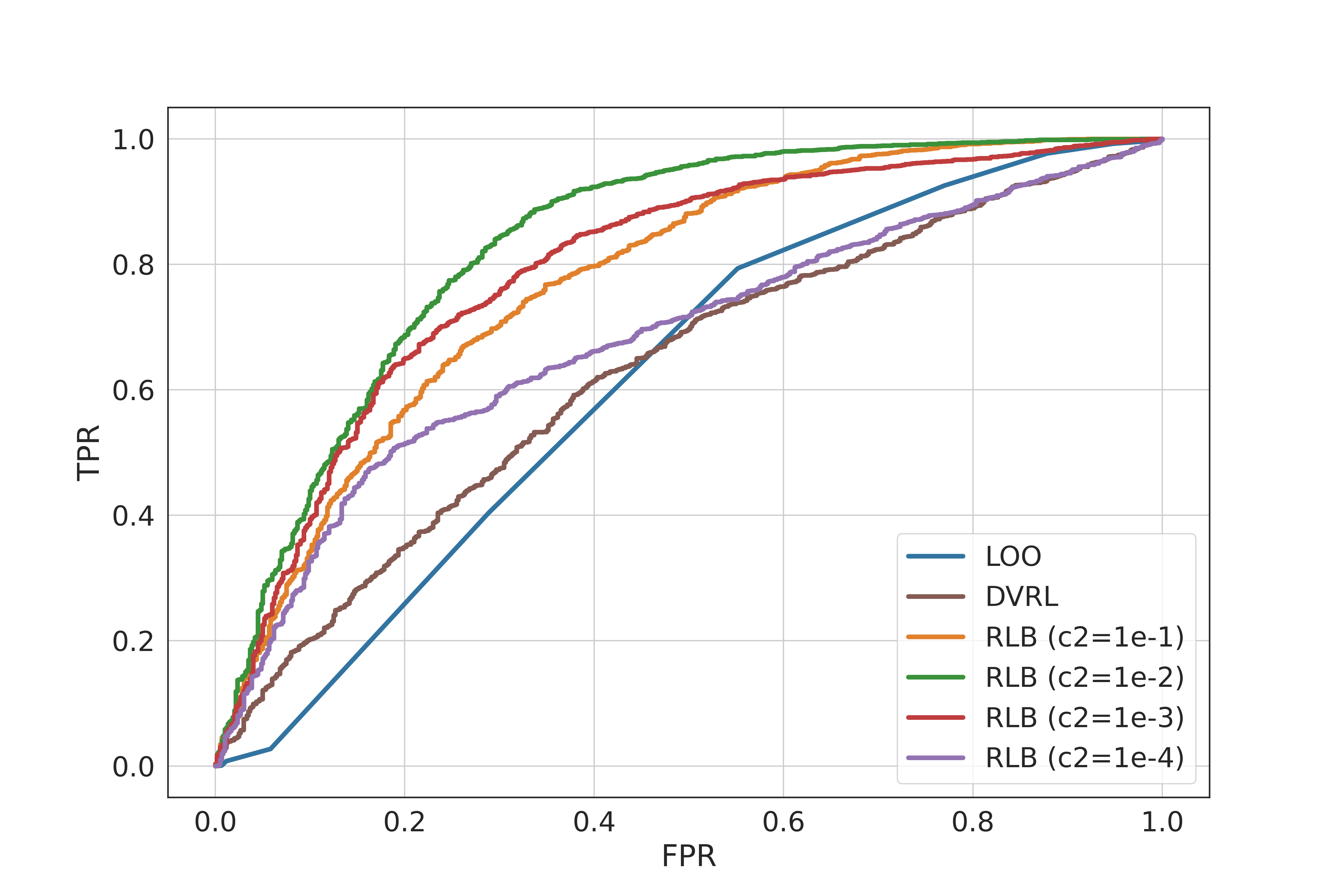}
            \caption{15\% error rate, best run in terms of AUC for each method.}
            \label{fig:roc_cifar10_15}
        \end{subfigure}
        \hfill
        \begin{subfigure}{0.65\textwidth}
            \centering
            \includegraphics[width=\textwidth]{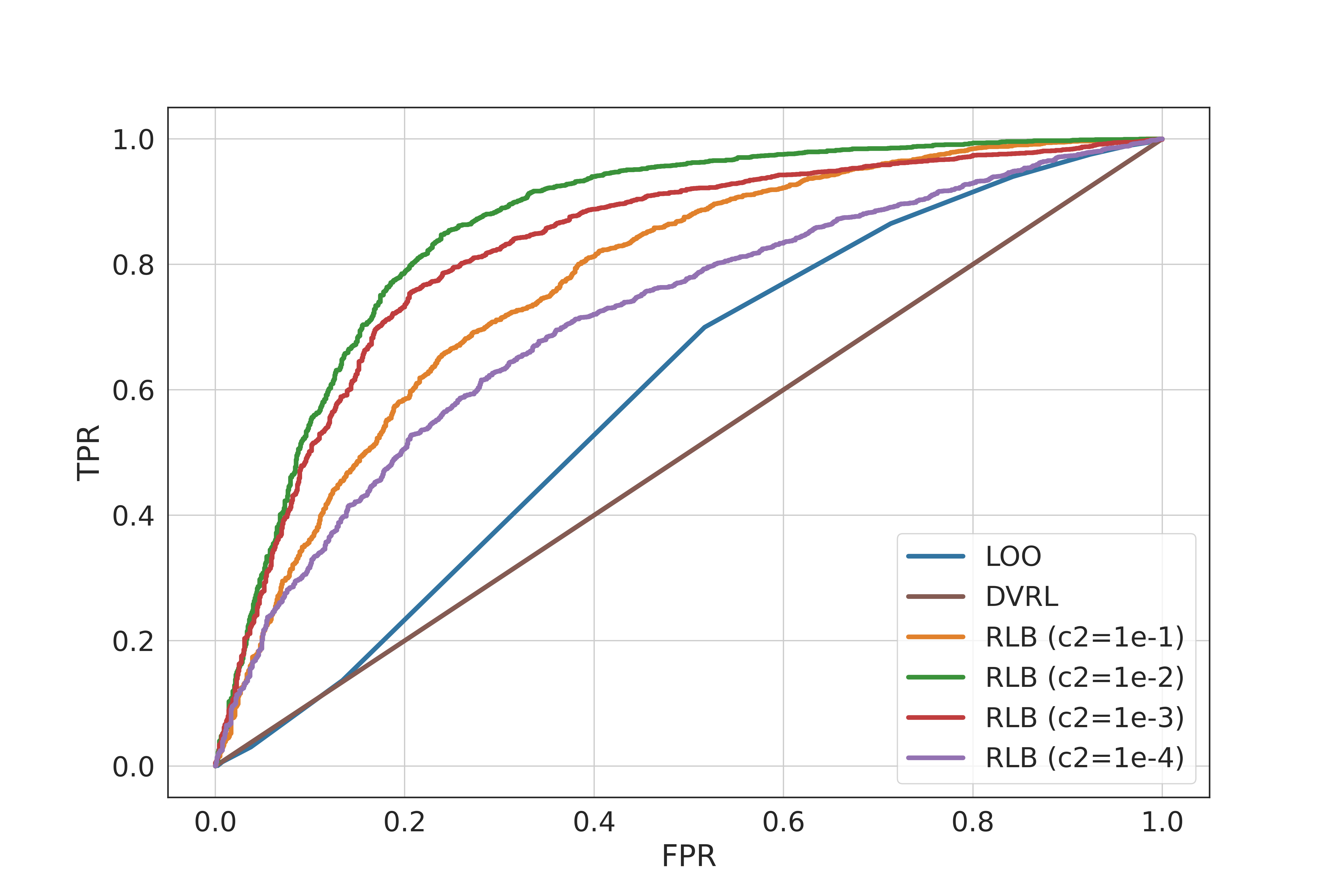}
            \caption{30\% error rate, best run in terms of AUC for each method.}
            \label{fig:roc_cifar10_3}
        \end{subfigure}
        \begin{subfigure}{0.65\textwidth}
            \centering
            \includegraphics[width=\textwidth]{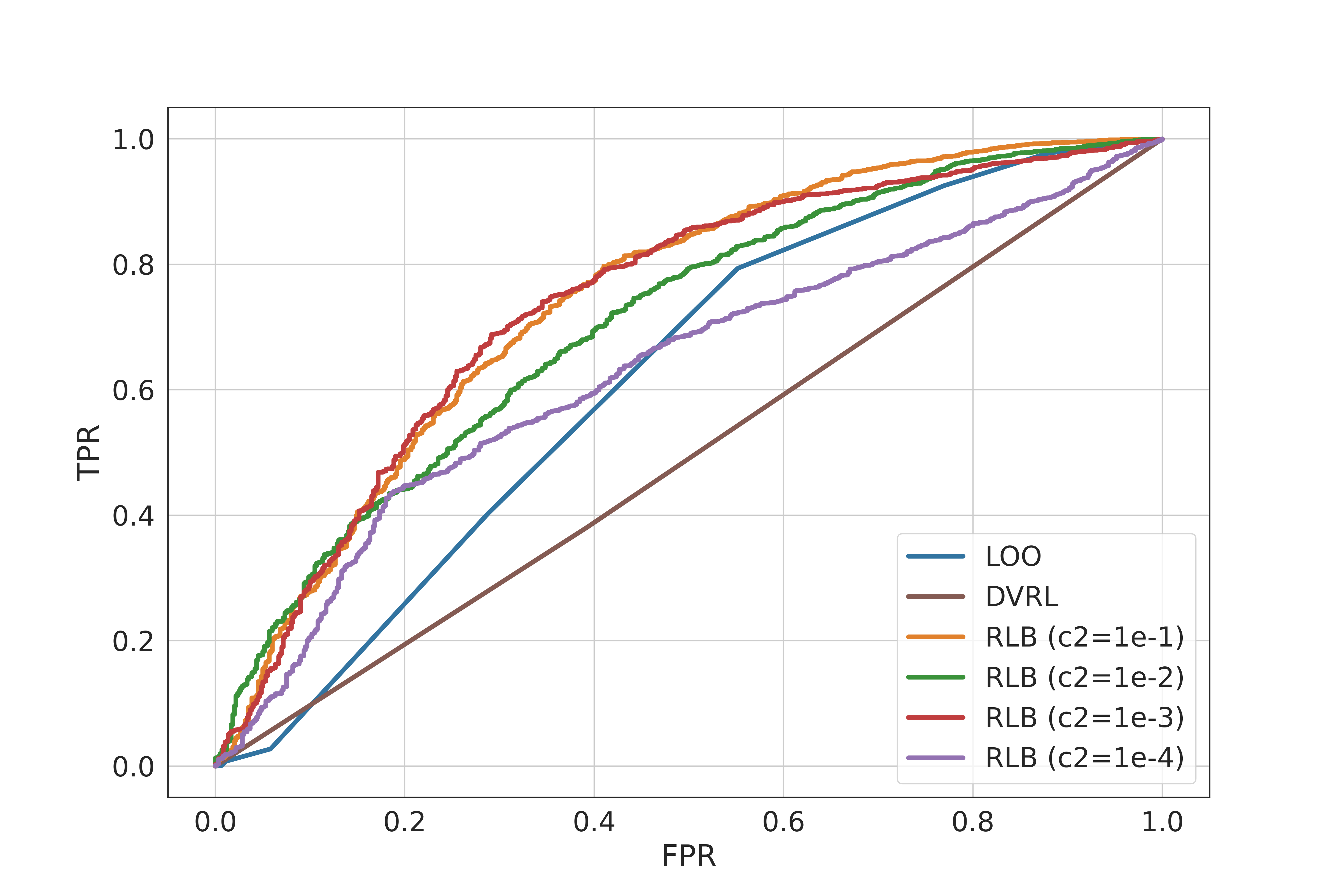}
            \caption{15\% error rate, worst run in terms of AUC for each method.}
            \label{fig:roc_cifar10_15_wrong}
        \end{subfigure}
        \hfill
        \begin{subfigure}{0.65\textwidth}
            \centering
            \includegraphics[width=\textwidth]{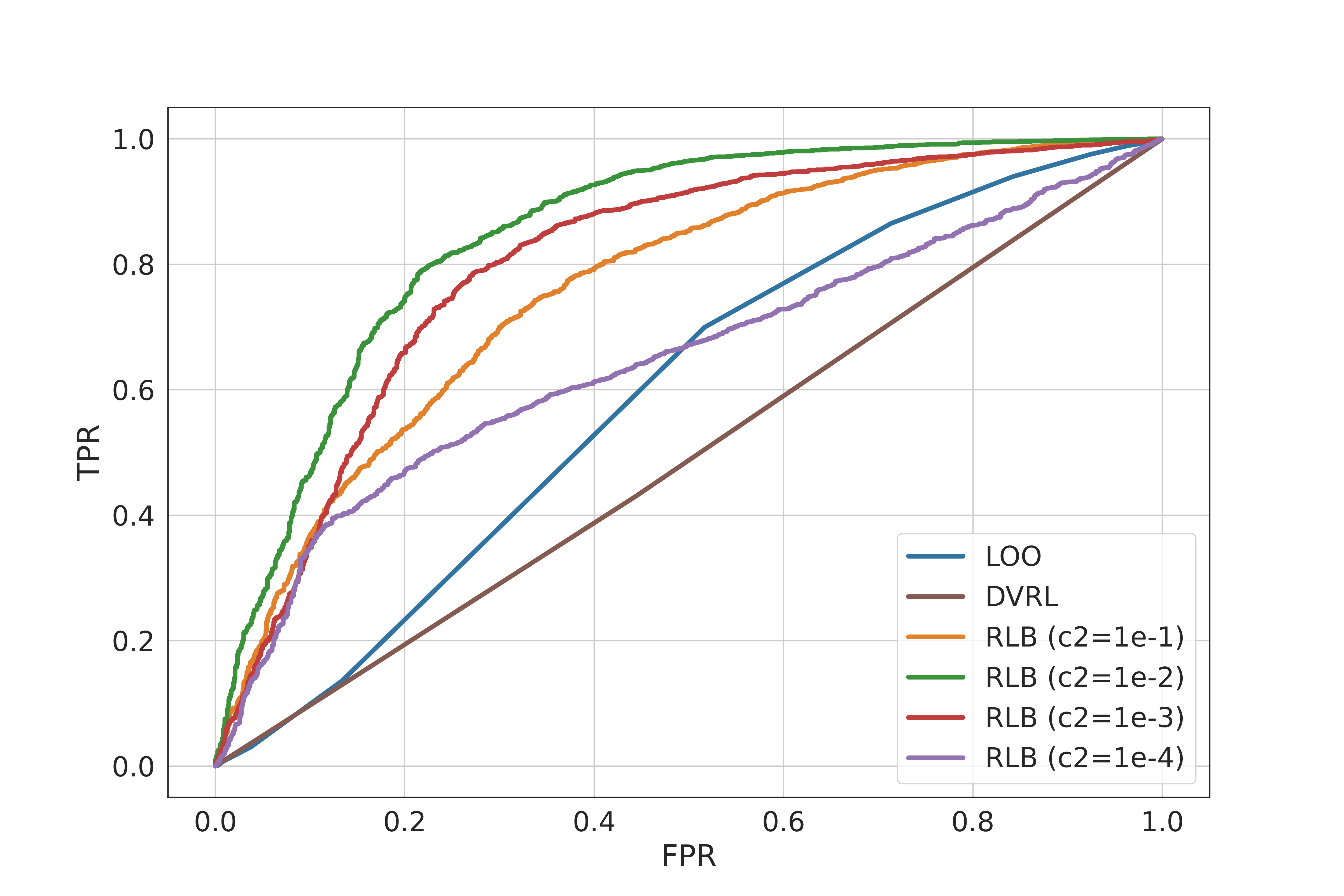}
            \caption{30\% error rate, worst run in terms of AUC for each method.}
            \label{fig:roc_cifar10_3_wrong}
        \end{subfigure} 
    \caption{ROC curves to measure the ability to detect noisy data in the CIFAR10 dataset, where the positive class is the data without noise and the negative class is the data with noise.}
    \label{fig:rocs_cifar10}
    \end{adjustwidth}
\end{figure}

It should be noted that in the case of CIFAR10 the results corroborate the quality of the agent and are quite promising as can be seen at Figures \ref{fig:tab_scores} and \ref{fig:rocs_cifar10}.

% ROCs

% Adult AUCs
\begin{table}[H]
    \footnotesize
    \centering
    \begin{tabular}{@{}l||ll@{}|}
                    & \textbf{15\% noise}               & \textbf{30\% noise}                \\\midrule
\textbf{LOO}        & $0.339 (\pm 0.0)$           & $0.509 (\pm 0.0)$            \\
\textbf{SHAP}       & $0.729 (\pm 0.018)$         & $0.724 (\pm 0.019)$          \\
\textbf{DVRL}       & $0.657 (\pm 0.193)$         & $0.734 (\pm 0.195)$          \\\midrule
\textbf{RLB (1e-1)} & $0.816 (\pm 0.006)$         & $\mathbf{0.856 (\pm 0.009)}$ \\
\textbf{RLB (1e-2)} & $\mathbf{0.831 (\pm 0.01)}$ & $0.825 (\pm 0.015)$          \\
\textbf{RLB (1e-3)} & $0.815 (\pm 0.009)$         & $0.839 (\pm 0.008)$          \\
\textbf{RLB (1e-4)} & $0.828 (\pm 0.008)$         & $0.819 (\pm 0.013)$          \\\bottomrule
\end{tabular}

    \caption{Tables of AUCs detecting noisy data by each of the models in the Adult dataset}
    \label{table:tab_aucs_adult}
\end{table}

% A5A AUCs
\begin{table}[H]
    \footnotesize
    \centering
    
\begin{tabular}{@{}l||ll@{}|}
                    & \textbf{15\% noise}                & \textbf{30\% noise}               \\\midrule
\textbf{LOO}        & $0.688 (\pm 0.0)$            & $0.535 (\pm 0.0)$           \\
\textbf{SHAP}       & -----                        & -----                       \\
\textbf{DVRL}       & $0.791 (\pm 0.141)$          & $0.68 (\pm 0.167)$          \\\midrule
\textbf{RLB (1e-1)} & $0.833 (\pm 0.016)$          & $0.847 (\pm 0.007)$         \\
\textbf{RLB (1e-2)} & $\mathbf{0.845 (\pm 0.007)}$ & $\mathbf{0.86 (\pm 0.005)}$ \\
\textbf{RLB (1e-3)} & $0.833 (\pm 0.008)$          & $0.845 (\pm 0.006)$         \\
\textbf{RLB (1e-4)} & $0.819 (\pm 0.007)$          & $0.832 (\pm 0.018)$         \\\bottomrule
\end{tabular}

    \caption{Tables of AUCs detecting noisy data by each of the models in the A5A dataset}
    \label{table:tab_aucs_a5a}
\end{table}

% W5A AUCs
\begin{table}[H]
    \footnotesize
    \centering
    \begin{tabular}{@{}l||ll@{}|}
                    & \textbf{15\% noise}                & \textbf{30\% noise}                \\\midrule
\textbf{LOO}        & $0.628 (\pm 0.0)$            & $0.455 (\pm 0.0)$            \\
\textbf{SHAP}       & -----                        & -----                        \\
\textbf{DVRL}       & $0.946 (\pm 0.008)$          & $0.859 (\pm 0.185)$          \\\midrule
\textbf{RLB (1e-1)} & $\mathbf{0.957 (\pm 0.013)}$ & $\mathbf{0.967 (\pm 0.003)}$ \\
\textbf{RLB (1e-2)} & $0.947 (\pm 0.008)$          & $0.967 (\pm 0.006)$          \\
\textbf{RLB (1e-3)} & $0.916 (\pm 0.008)$          & $0.933 (\pm 0.006)$          \\
\textbf{RLB (1e-4)} & $0.887 (\pm 0.007)$          & $0.892 (\pm 0.025)$          \\\bottomrule
\end{tabular}

    \caption{Tables of AUCs detecting noisy data by each of the models in the W5A dataset}
    \label{table:tab_aucs_W5a}
\end{table}

% CIFAR10 AUCs
\begin{table}[H]
    \footnotesize
    \centering
    
\begin{tabular}{@{}l||ll@{}|}
                    & \textbf{15\% noise}                & \textbf{30\% noise}                \\\midrule
\textbf{LOO}        & $0.619 (\pm 0.0)$            & $0.592 (\pm 0.0)$            \\
\textbf{SHAP}       & -----                        & -----                        \\
\textbf{DVRL}       & $0.548 (\pm 0.061)$          & $0.498 (\pm 0.003)$          \\\midrule
\textbf{RLB (1e-1)} & $0.757 (\pm 0.014)$          & $0.767 (\pm 0.005)$          \\
\textbf{RLB (1e-2)} & $\mathbf{0.779 (\pm 0.044)}$ & $\mathbf{0.855 (\pm 0.005)}$ \\
\textbf{RLB (1e-3)} & $0.764 (\pm 0.019)$          & $0.82 (\pm 0.009)$           \\
\textbf{RLB (1e-4)} & $0.657 (\pm 0.021)$          & $0.678 (\pm 0.021)$          \\\bottomrule
\end{tabular}

    \caption{Tables of AUCs detecting noisy data by each of the models in the binarized CIFAR10 dataset}
    \label{table:tab_aucs_cifar}
\end{table}

In Appendices~\ref{appendix:a5a} and \ref{appendix:w5a} we report the result graphs of the rest of the ROC analysis on the proposed tabular datasets.

\subsection{Image data}
As previously mentioned in Subsection~\ref{subsec:arq}, this algorithm is not restricted only to its application to tabular data. Any problem in which a data can be vectorized in advance can be used to evaluate the quality of the given samples. 

\begin{table}[H]
    \centering
    \caption{Details of the records, features and classes of the image dataset to be used in the experimentation.}
    \begin{tabular}{c|ccc|cc}
    \toprule
    \textbf{Dataset} & \textbf{Train} & \textbf{Validation} & \textbf{Test} & \textbf{\# Features} & \textbf{\# Classes} \\ \midrule
    MNIST & 5k & 1k & 5k & 1024 & 10 \\  \bottomrule
    \end{tabular}
    \label{table:image_dataset}
\end{table}

In this case it has been decided to test the robustness of the model by evaluating the quality of vectorized image data using the MNIST dataset~\citep{deng2012mnist}. The way to vectorize these images has been through a CLIP model~\citep{Radford2021LearningSupervision} with a pre-trained ResNet50. At Table~\ref{table:image_dataset} can be seen the different sets and the specification of each one after vectorizing and splitting the data.

Since this dataset contains handwritten digits from 0 to 9, it was also decided to complicate a little more the error introduced to the data. Thus, each of the erroneous data has been classified as the class of the next digit in a circular fashion, so that 0 becomes 1, 1 becomes 2, etc. and 9 becomes 0. 

\begin{table}
    \footnotesize
    \centering
    \caption{Table of scores against test data using the final filtering proposed by each of the models in the MNIST data set}
    
\begin{tabular}{@{}l||lll@{}|}
                                        & \textbf{0\% noise}      & \textbf{15\% noise}     & \textbf{30\% noise}                        \\ \midrule
\textbf{Baseline}                       & $\mathbf{0.968}$  & $0.779$           & $0.621$                              \\
\textbf{LOO}                            & $0.968(\pm0.000)$ & $0.780(\pm0.000)$ & $0.620(\pm0.000)$                    \\
\textbf{SHAP}                           & -                 & -                 & -                                    \\
\textbf{DVRL}                           & $0.908(\pm0.119)$ & $0.923(\pm0.081)$ & $0.780(\pm0.379)$                    \\ \midrule
\textbf{RLB (1e-1)}                     & $0.967(\pm0.001)$ & $0.890(\pm0.013)$ & $0.782(\pm0.017)$                    \\
\textbf{RLB (1e-2)}                     & $0.968(\pm0.002)$ & $0.910(\pm0.007)$ & $0.842(\pm0.016)$                    \\
\textbf{RLB (1e-3)}                     & $0.969(\pm0.002)$ & $\mathbf{0.914(\pm0.009)}$ & $\mathbf{0.867(\pm0.018)}$  \\
\textbf{RLB (1e-4)}                     & $0.968(\pm0.001)$ & $0.876(\pm0.011)$ & $0.802(\pm0.030)$                    \\ \bottomrule
\end{tabular}
\label{tab:scores_mnist}

    \label{table:img_scores}
\end{table}

% Scores
\begin{figure}[H]
\begin{adjustwidth}{-1.5cm}{-1.5cm}
\centering
\includegraphics[width=\textwidth]{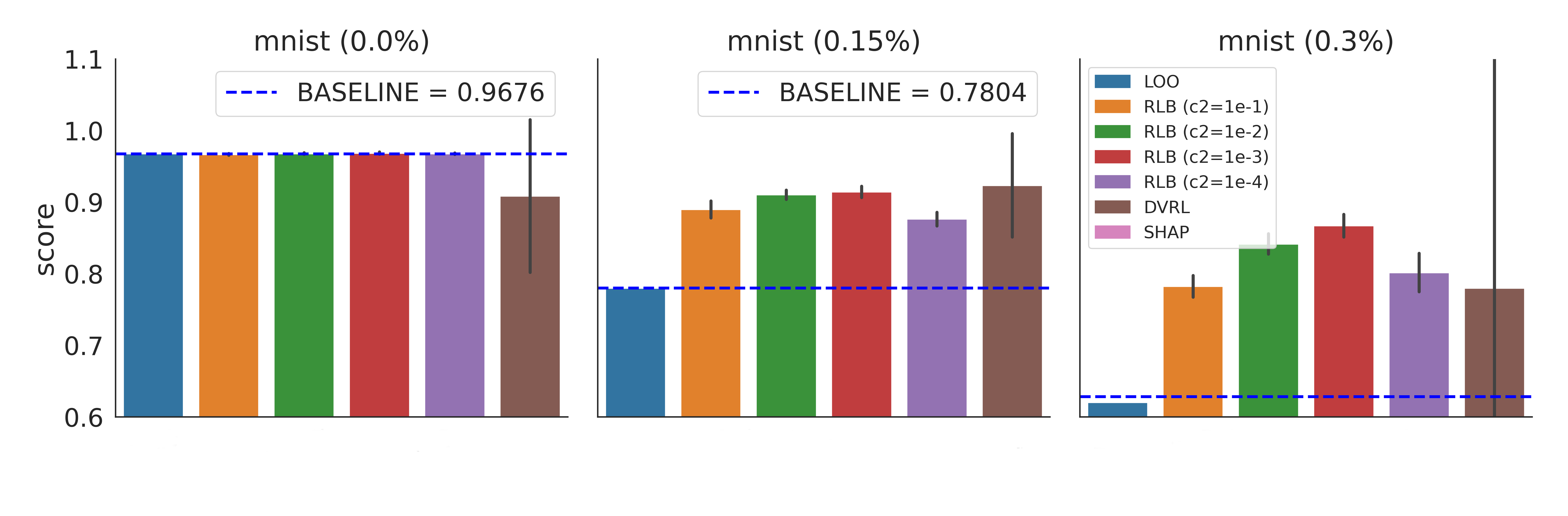}
\caption{Final scores of the agents trained over the MNIST data}
\label{fig:mnist_scores}
\end{adjustwidth}
\end{figure}

% ROCs
\begin{figure}[H]
    \begin{adjustwidth}{-2cm}{-2cm}
        \centering
        \begin{subfigure}{0.65\textwidth}
            \centering
            \includegraphics[width=\textwidth]{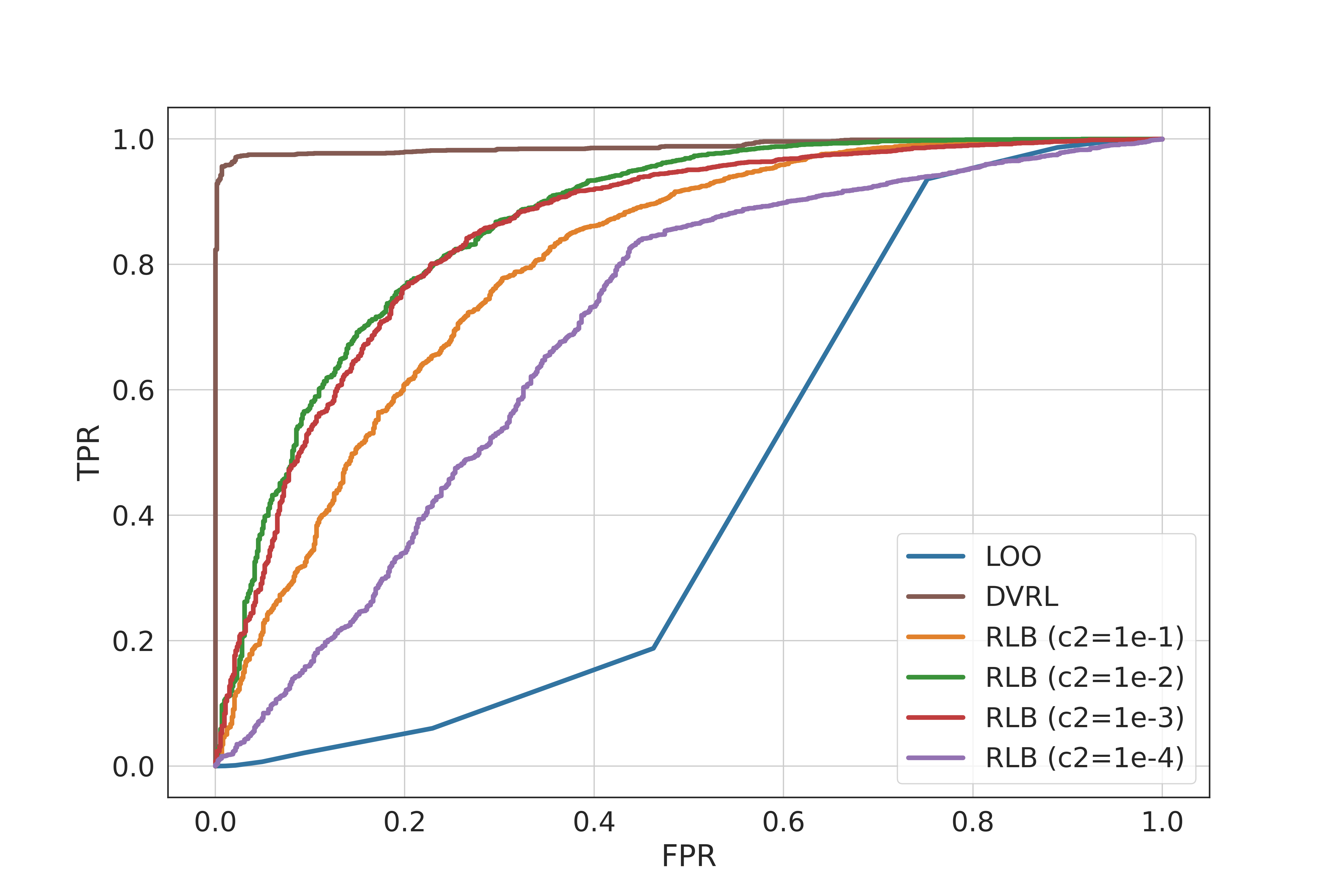}
            \caption{15\% error rate, best run in terms of AUC for each method.}
            \label{fig:roc_mnist_15}
        \end{subfigure}
        \hfill
        \begin{subfigure}{0.65\textwidth}
            \centering
            \includegraphics[width=\textwidth]{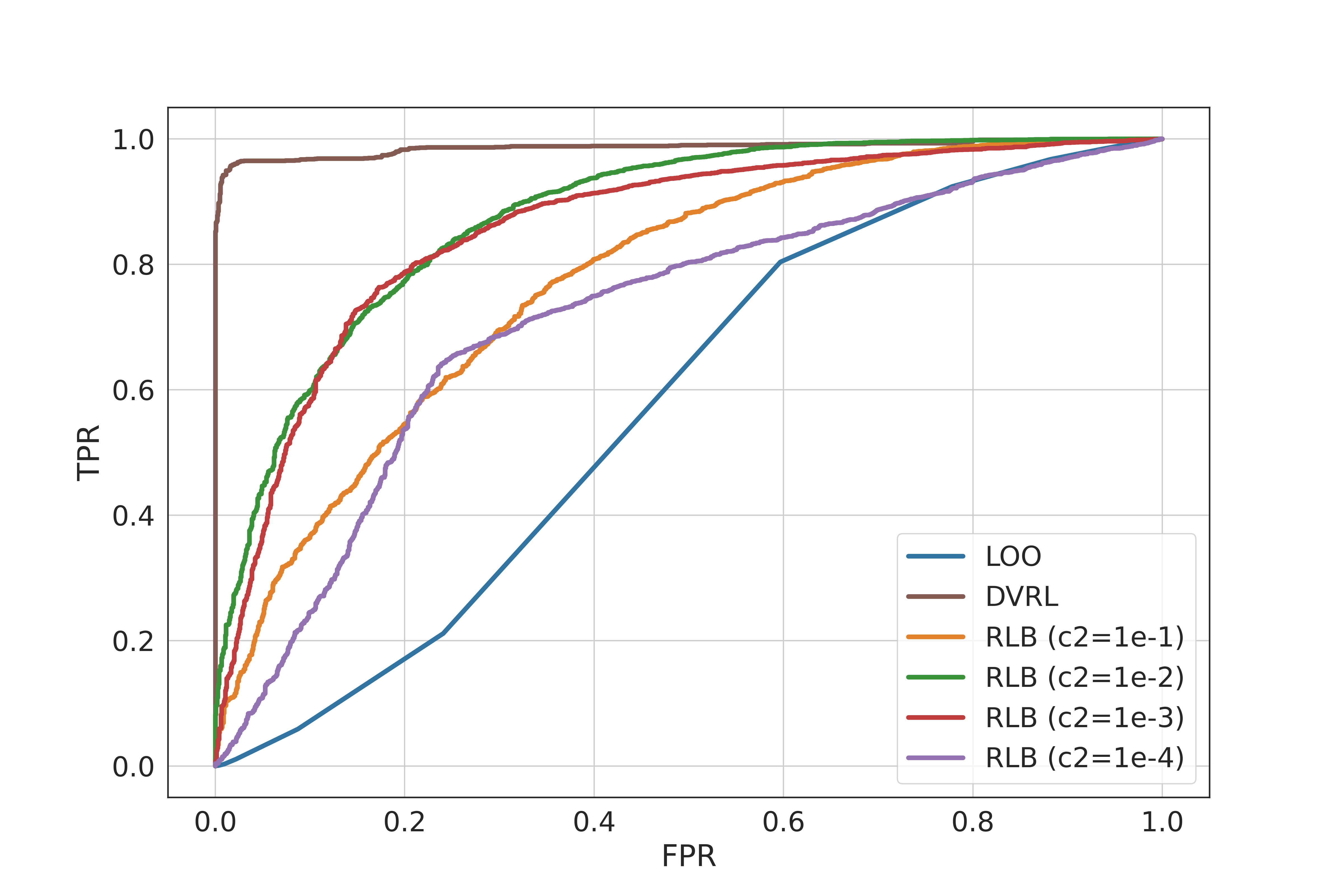}
            \caption{30\% error rate, best run in terms of AUC for each method.}
            \label{fig:roc_mnist_3}
        \end{subfigure}
        \begin{subfigure}{0.65\textwidth}
            \centering
            \includegraphics[width=\textwidth]{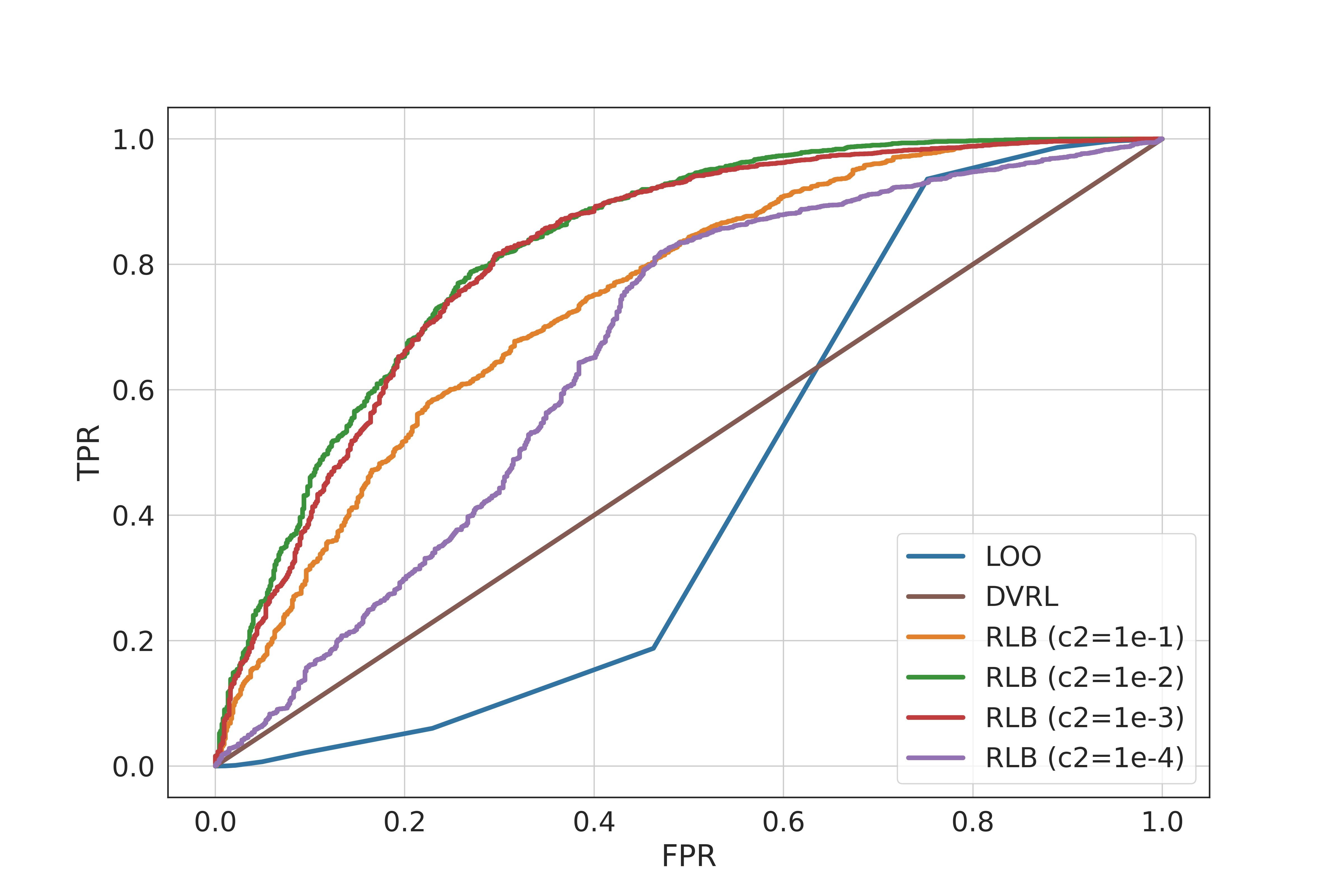}
            \caption{15\% error rate, worst run in terms of AUC for each method.}
            \label{fig:roc_mnist_15_wrong}
        \end{subfigure}
        \hfill
        \begin{subfigure}{0.65\textwidth}
            \centering
            \includegraphics[width=\textwidth]{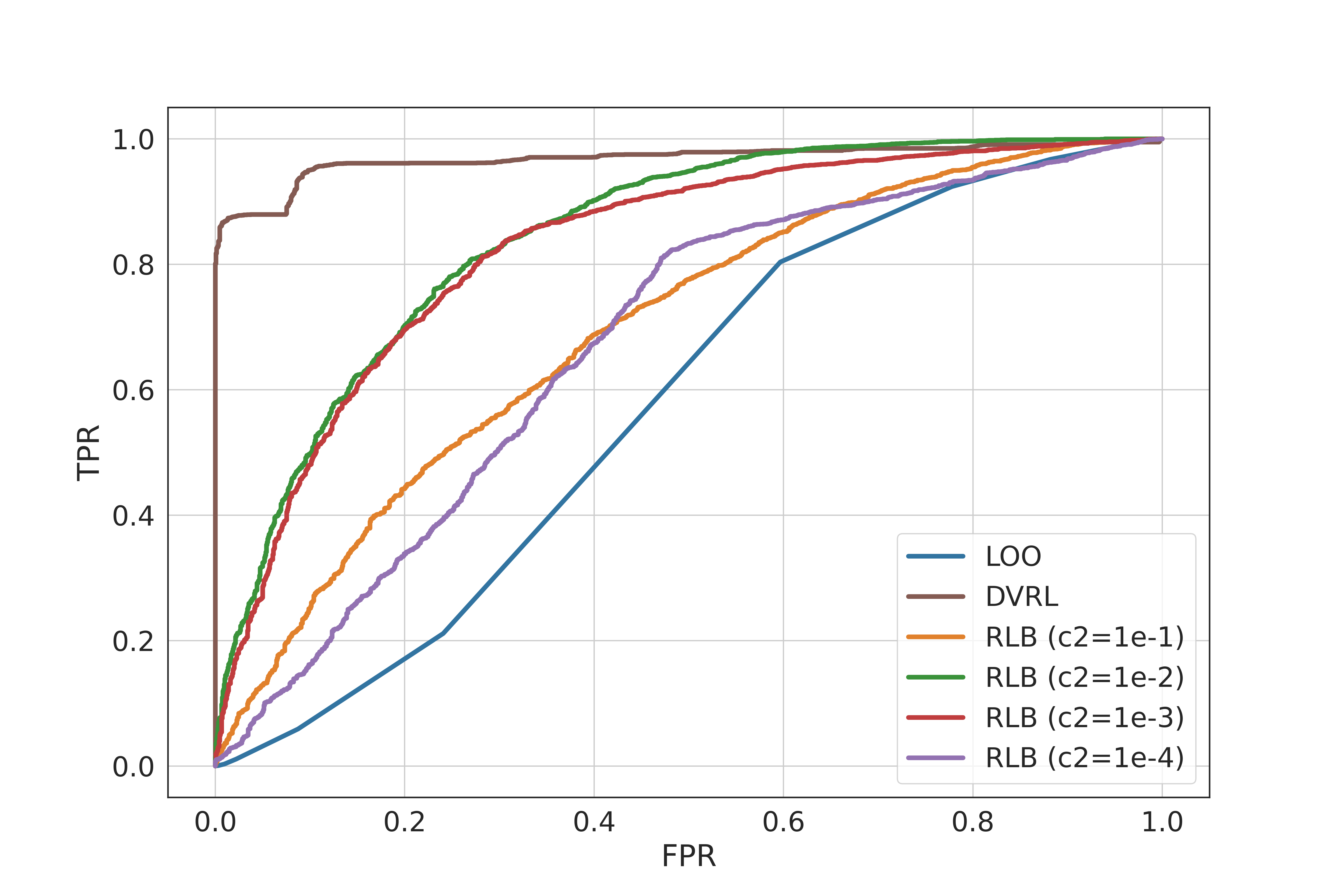}
            \caption{30\% error rate, worst run in terms of AUC for each method.}
            \label{fig:roc_mnist_3_wrong}
        \end{subfigure} 
    \caption{ROC curves to measure the ability to detect noisy data in the MNIST dataset, where the positive class is the data without noise and the negative class is the data with noise.}
    \label{fig:rocs_mnist}
    \end{adjustwidth}
\end{figure}

\begin{table}
\footnotesize
\centering
\caption{Table of AUCs detecting noisy data by each of the models in image dataset}
\begin{tabular}{@{}l||ll@{}|}
                    & \textbf{15\% noise}              & \textbf{30\% noise}              \\\midrule
\textbf{LOO}        & $0.440(\pm0.0)$            & $0.575(\pm0.0)$            \\
\textbf{SHAP}       & -                          & -                          \\
\textbf{DVRL}       & $0.884(\pm0.192)$          & $\mathbf{0.977(\pm0.006)}$ \\\midrule
\textbf{RLB (1e-1)} & $0.816 (\pm 0.006)$        & $0.737(\pm0.03)$           \\
\textbf{RLB (1e-2)} & $\mathbf{0.853(\pm0.014)}$ & $0.861(\pm0.012)$          \\
\textbf{RLB (1e-3)} & $0.839(\pm0.014)$          & $0.843(\pm0.012)$          \\
\textbf{RLB (1e-4)} & $0.679(\pm0.013)$          & $0.688(\pm0.016)$          \\\bottomrule
\end{tabular}

\label{table:tab_aucs_mnist}
\end{table}

In view of the above results, we can conclude that the RLBoost method is really effective for its purpose. In each of the tests performed against the proposed dataset, it has ended up with, if not the best value in terms of accuracy after agent filtering, a fairly competitive one. One of the striking cases of these results is the exceptional good performance of the DVRL method in the case where the noise is $30\%$, in detection of noisy data in Table~\ref{table:tab_aucs_mnist}. However, in view of the results of other runs, we can see that DVRL suffers from instability when it comes to yield good results, while the proposed method does not.

\section{Conclusions and future work}
\label{sec:concl}
The problem of data evaluation can be oriented as if it were a search problem, and in particular it is possible to formulate it as a trajectory free reinforcement learning problem. This approach makes it possible to understand that each data point is part of a context and that it will be more or less valuable depending on the supervised learning model to be used to model it.

This paper opens a new framework in the evaluation of data from a supervised learning model using reinforcement learning without any trajectory approach. In this sense, it opens up to us different branches of research.
First, one of the next steps is to apply this strategy to text classification problems in order to be able to extend a training dataset, since this problem usually has the difficulty that the manual evaluation of records by is often very expensive and specialized. Therefore, a properly labeled validation set can be developed with the goal of collecting training data automatically for further evaluation.
On the other hand, it also opens the possibility of improving the reinforcement learning algorithm (perhaps using SAC~\citep{pmlr-v80-haarnoja18b} or V-MPO~\citep{Song2020V-MPO}), taking into account that it must be of the Actor-Critic type. This improvement should have much emphasis on the sample-efficiency of the method, since this will allow a faster convergence to more complex estimators.
Finally, there could be some improvements at the model architecture, like using LongFormer from~\citet{LongFormer} to try larger batches without increasing the computational cost of the algorithm, or using a Encoder/Decoder architecture for the policy network.
\begin{appendices} \label{appendixes}

\section{A5A dataset} \label{appendix:a5a}

% ROCs
\begin{figure}[H]
    \begin{adjustwidth}{-2cm}{-2cm}
        \centering
        \begin{subfigure}{0.65\textwidth}
            \centering
            \includegraphics[width=\textwidth]{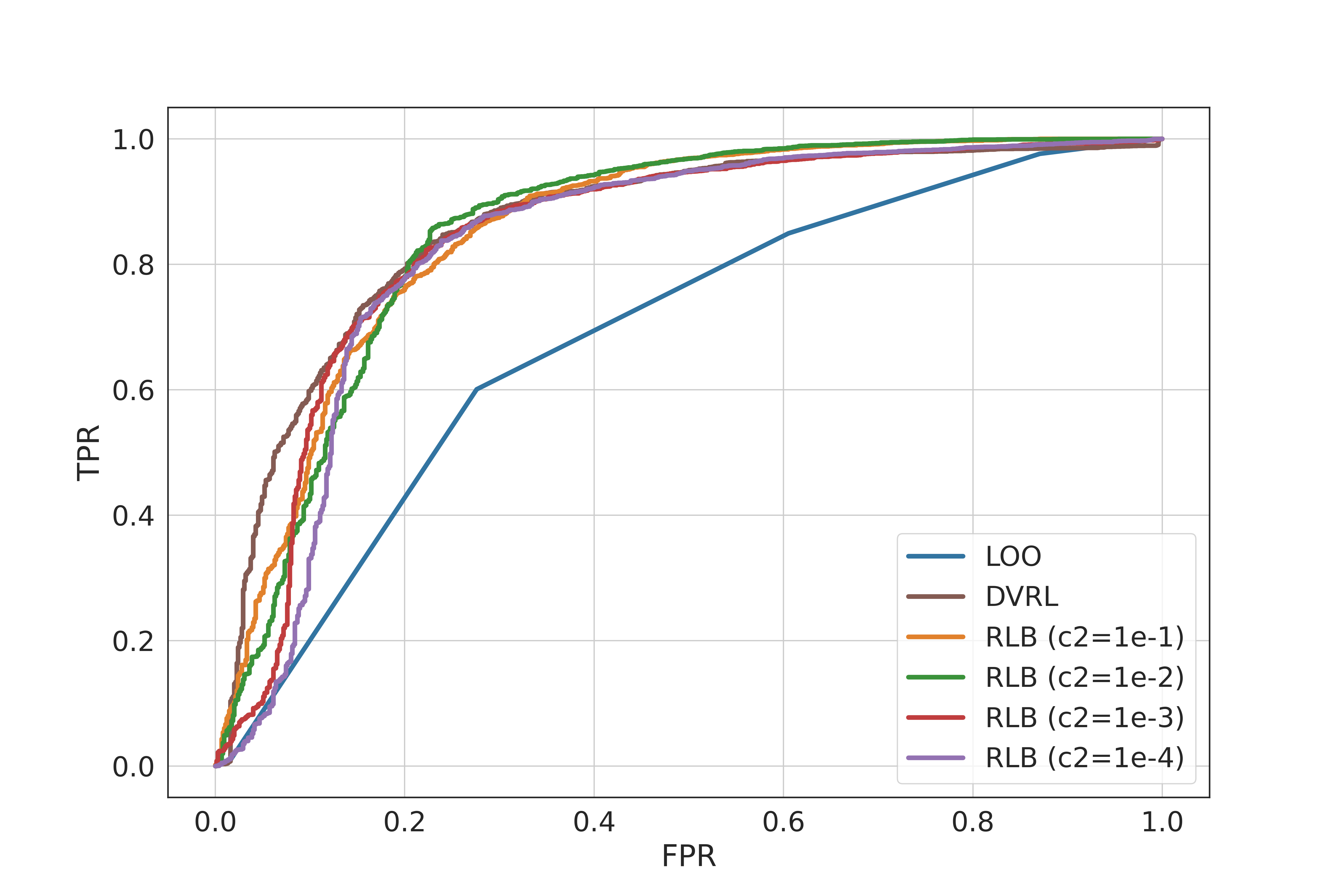}
            \caption{ROC curve of the different models on the data set with 15\% error rate.}
            \label{fig:roc_a5a_15}
        \end{subfigure}
        \hfill
        \begin{subfigure}{0.65\textwidth}
            \centering
            \includegraphics[width=\textwidth]{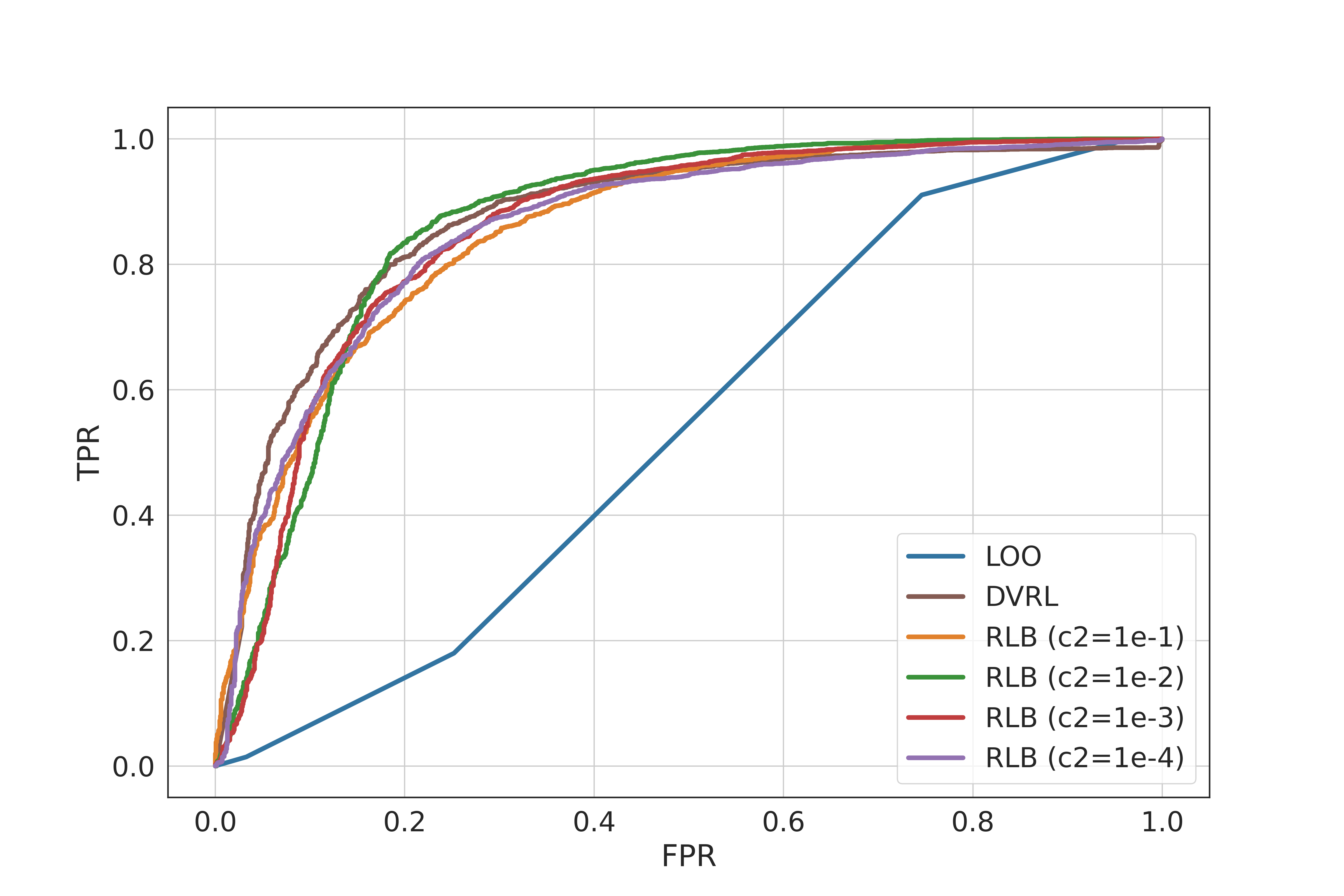}
            \caption{ROC curve of the different models on the data set with 30\% error rate.}
            \label{fig:roc_a5a_3}
        \end{subfigure} 
    \caption{ROC curves to measure the ability to detect noisy data in the A5A dataset, where the positive class is the data without noise and the negative class is the data without noise.}
    \label{fig:rocs_a5a}
    \end{adjustwidth}
\end{figure}

\section{W5A dataset} \label{appendix:w5a}

% ROCs
\begin{figure}[H]
    \begin{adjustwidth}{-2cm}{-2cm}
        \centering
        \begin{subfigure}{0.65\textwidth}
            \centering
            \includegraphics[width=\textwidth]{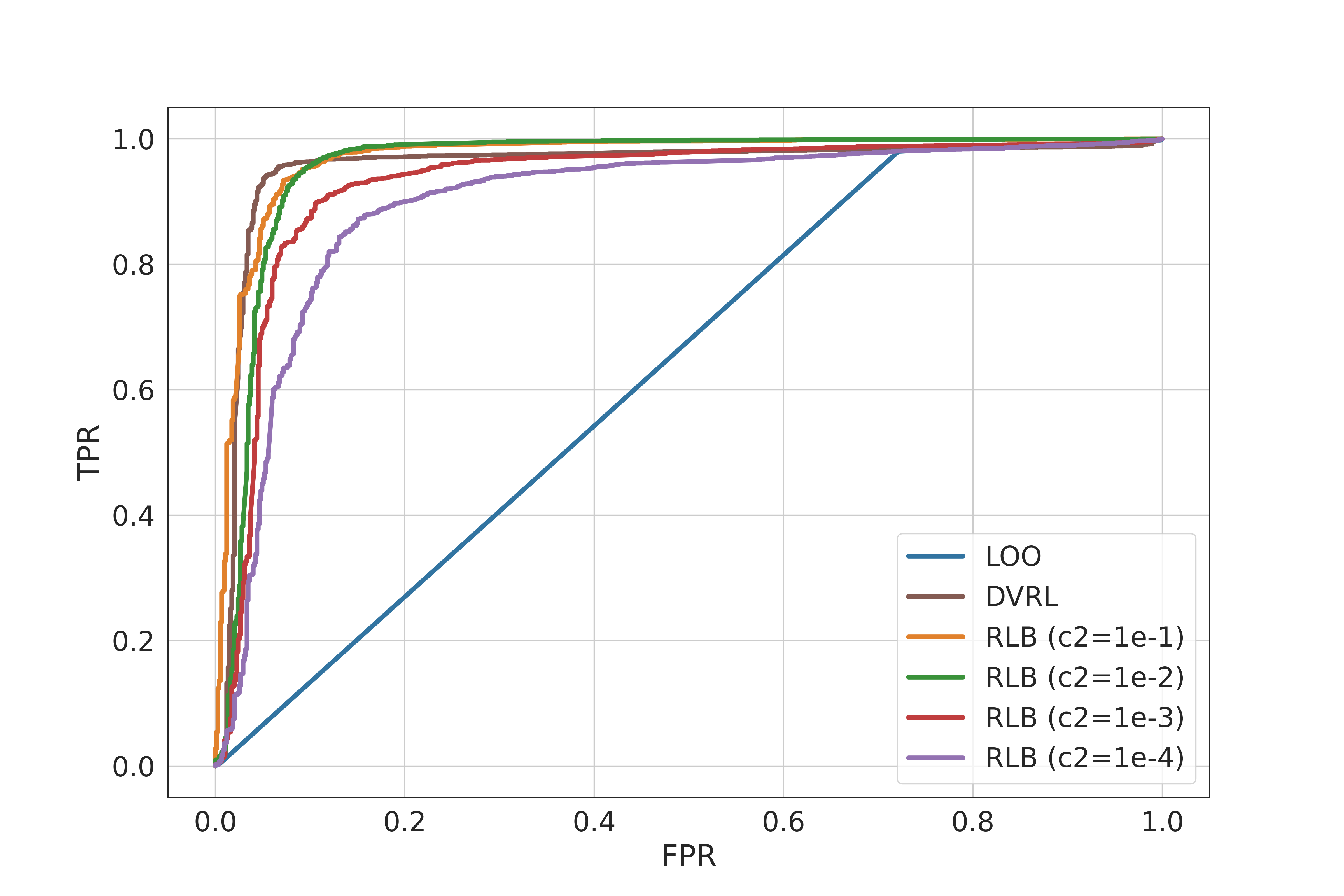}
            \caption{ROC curve of the different models on the data set with 15\% error rate.}
            \label{fig:roc_w5a_15}
        \end{subfigure}
        \hfill
        \begin{subfigure}{0.65\textwidth}
            \centering
            \includegraphics[width=\textwidth]{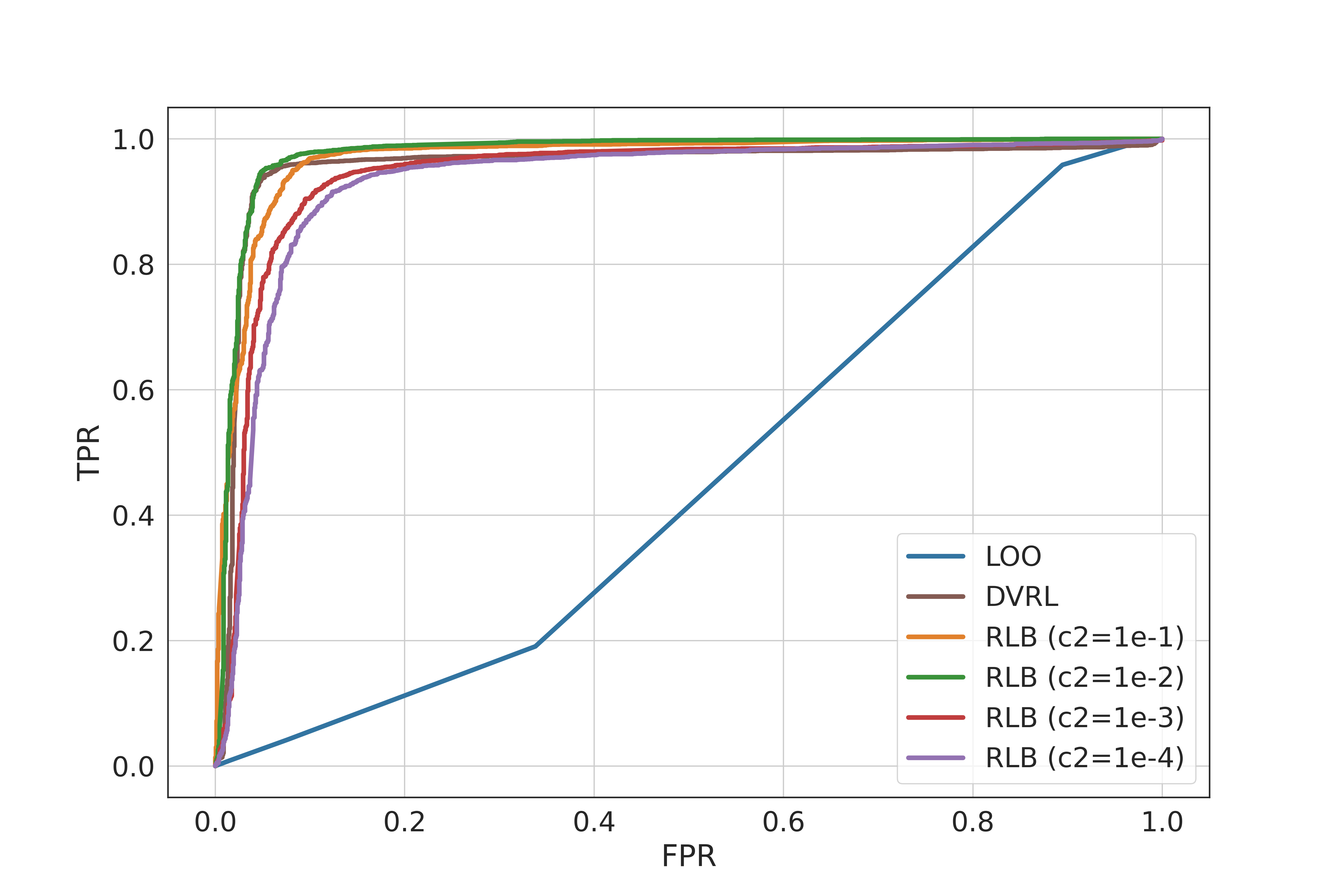}
            \caption{ROC curve of the different models on the data set with 30\% error rate.}
            \label{fig:roc_w5a_3}
        \end{subfigure} 
    \caption{ROC curves to measure the ability to detect noisy data in the W5A dataset, where the positive class is the data without noise and the negative class is the data without noise.}
    \label{fig:rocs_w5a}
    \end{adjustwidth}
\end{figure}

\section{Ablation study}\label{apendix:ablation}

To understand what elements or parts make ReSuLT work in this way, it has been decided to ablate different parts to see how it behaves towards them. In particular, it has been decided to see what influence the CLS token and the Transformer Encoder in general can have on the final performance of the agents. 

Therefore, we have executed 5 times each of the combinations shown previously in the experimental part but leaving only the baseline score as input value to the value function (SB), leaving only the CLS token derived from the Transformer Encoder with all the batch data (CLS) and leaving both values as input to the value function (CLS+SB).

\subsection{Adult ablation scores}

% Scores
\begin{figure}[H]
    \begin{adjustwidth}{-1.5cm}{-1.5cm}
    \centering
        \begin{subfigure}{0.4\textwidth}
            \centering
            \includegraphics[width=\textwidth]{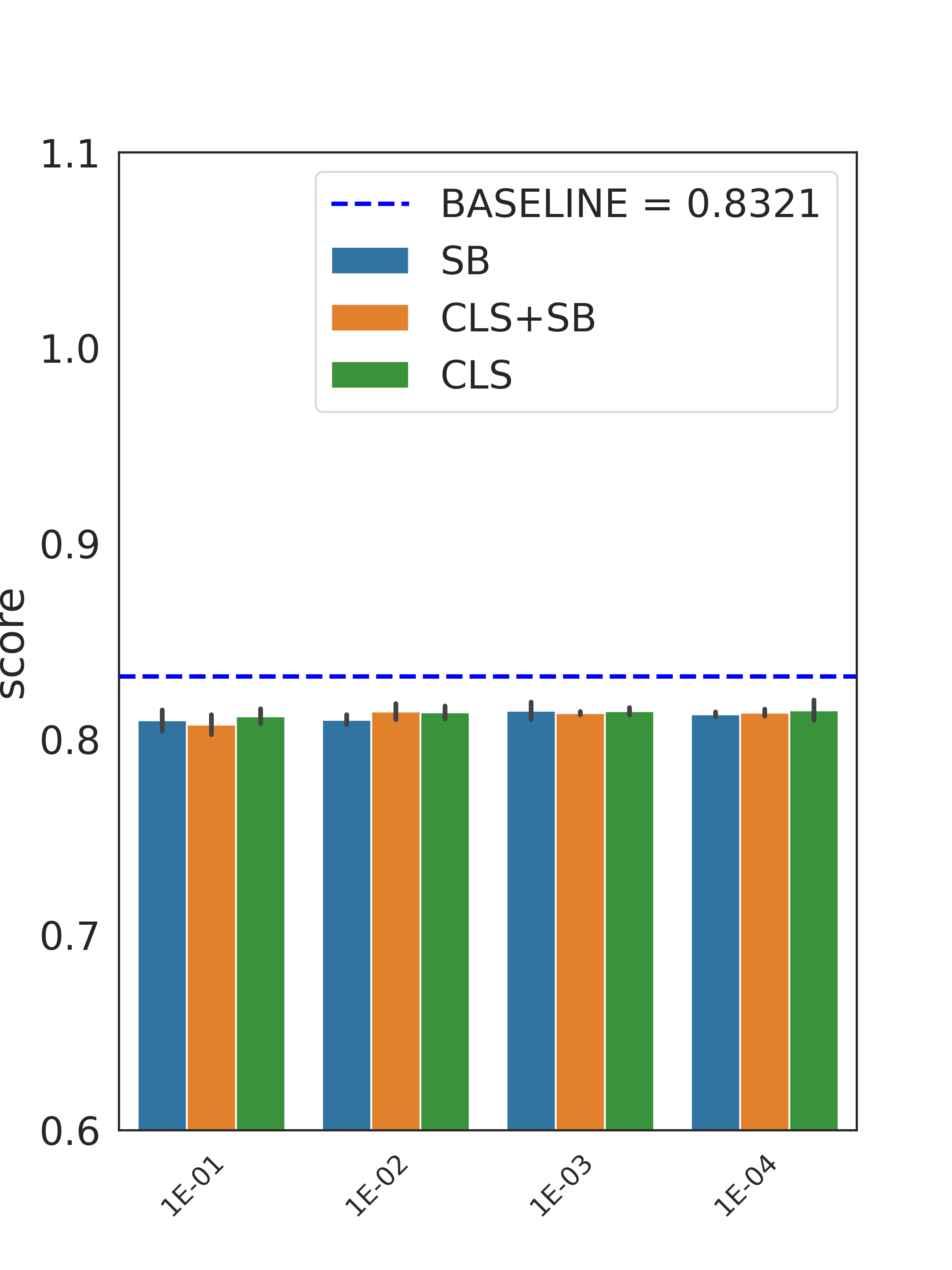}
            \caption{0\% noise}
            \label{fig:adult_score_0_ablation}
        \end{subfigure}\hfill
        \begin{subfigure}{0.4\textwidth}
            \centering
            \includegraphics[width=\textwidth]{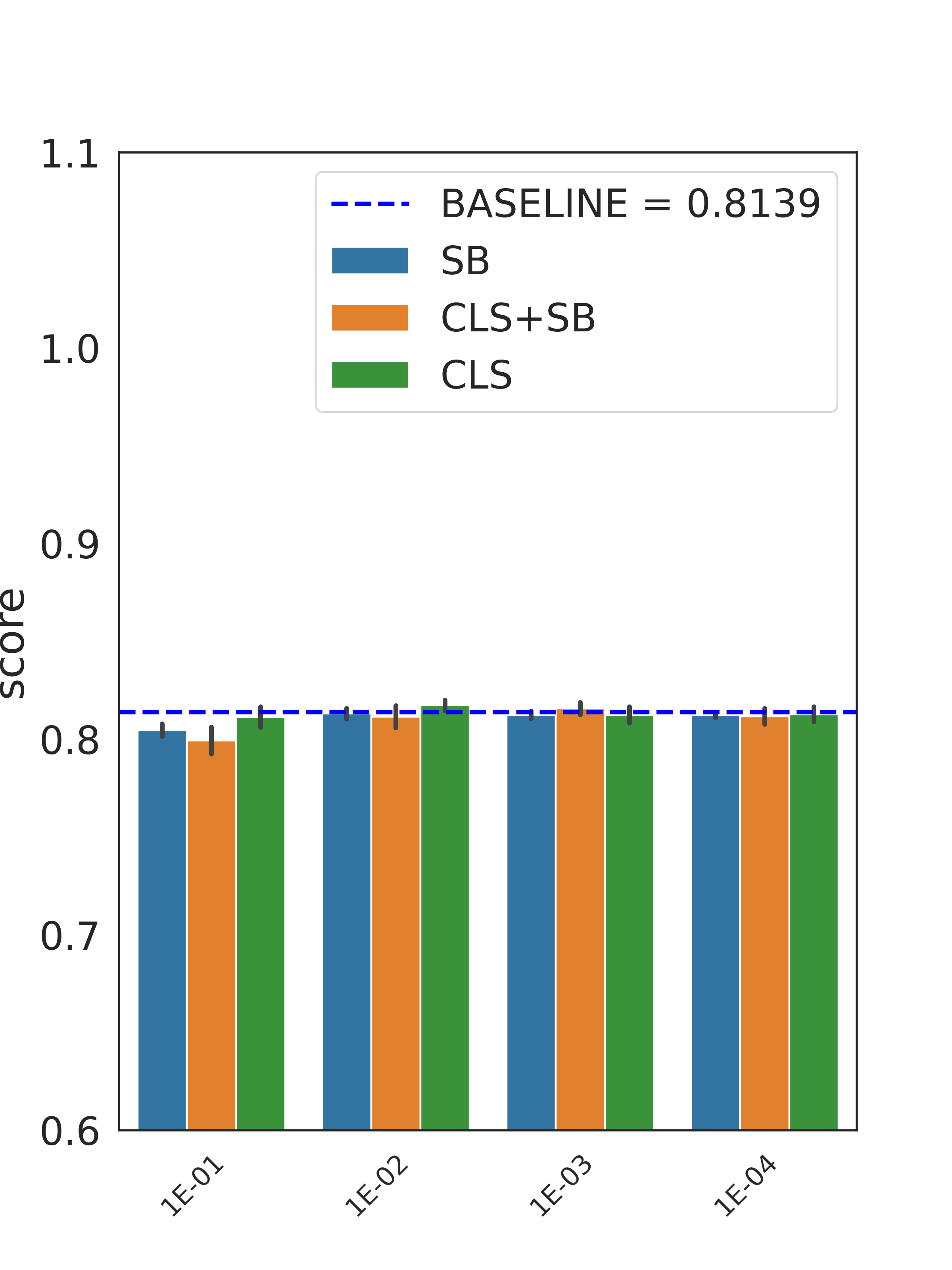}
            \caption{15\% noise}
            \label{fig:adult_score_15_ablation}
        \end{subfigure}\hfill
        \begin{subfigure}{0.4\textwidth}
            \centering
            \includegraphics[width=\textwidth]{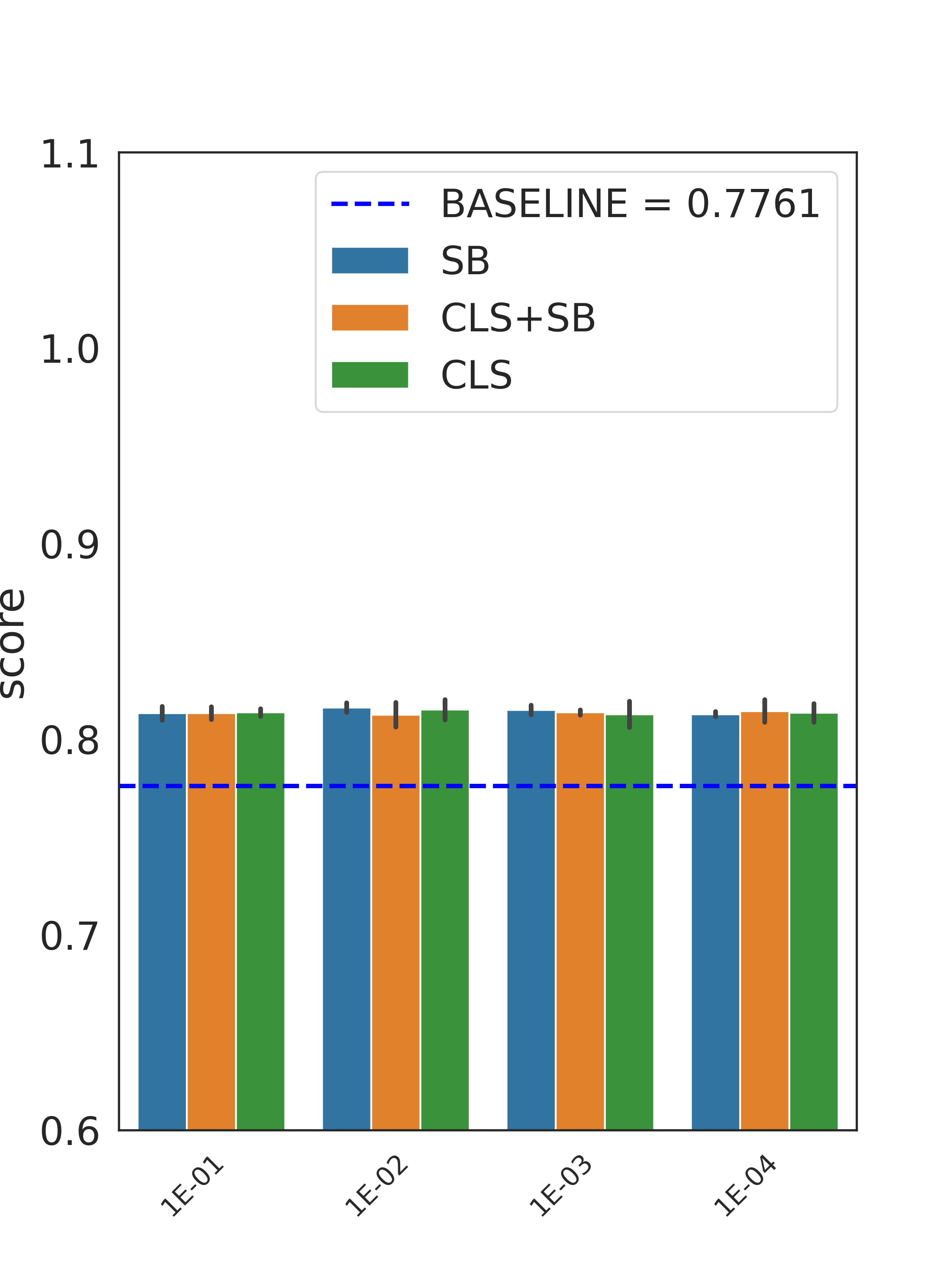}
            \caption{30\% noise}
            \label{fig:adult_score_3_ablation}
        \end{subfigure}
    \caption{Final Scores for the agents trained over the Adult dataset making the ablations}
    \label{fig:adult_scores_ablations}
    \end{adjustwidth}
\end{figure}

\begin{table}[H]
    \centering
    \footnotesize
    \begin{adjustwidth}{-1.5cm}{-1.5cm}
    \begin{tabular}{c|l|l||l|l|l|l}
    \hline
    \textbf{Data} & \multicolumn{1}{c|}{\textbf{Noise}} & \diagbox{\textbf{Method}}{\textbf{C2}}            & \multicolumn{1}{c|}{\textbf{1e-1}}      & \multicolumn{1}{c|}{\textbf{1e-2}}               & \multicolumn{1}{c|}{\textbf{1e-3}}               & \multicolumn{1}{c}{\textbf{1e-4}} \\ \hline
    \multirow{9}{*}{Adult} & {\multirow{3}{*}{0}}                       & SB     & $0.810(\pm0.006)$  & $0.810(\pm0.003)$           & $0.814(\pm0.005)$           & $0.813(\pm0.001)$ \\  
        &                                                               & CLS    & $0.812(\pm0.004)$  & $0.814(\pm0.004)$           & $\mathbf{0.814(\pm0.002)}$  & $0.815(\pm0.006)$ \\  
        &                                                               & CLS+SB & $0.807(\pm0.006)$  & $0.814(\pm0.005)$           & $0.813(\pm0.001)$           & $0.814(\pm0.002)$ \\ \cline{2-7} 
        
        & \multirow{3}{*}{15}                                           & SB     & $0.805(\pm0.004)$  &  $0.813(\pm0.003)$          &  $0.812(\pm0.002)$          & $0.812(\pm0.002)$ \\  
        &                                                               & CLS    & $0.811(\pm0.006)$  &  $\mathbf{0.817(\pm0.003)}$ &  $0.812(\pm0.004)$          & $0.813(\pm0.004)$ \\  
        &                                                               & CLS+SB & $0.799(\pm0.007)$  &  $0.811(\pm0.006)$          &  $0.816(\pm0.004)$          & $0.812(\pm0.004)$ \\ \cline{2-7} 
        
        & \multirow{3}{*}{30}                                           & SB     & $0.813(\pm0.004)$  &  $\mathbf{0.816(\pm0.003)}$ &  $0.815(\pm0.003)$          & $0.813(\pm0.001)$ \\  
        &                                                               & CLS    & $0.814(\pm0.002)$  &  $0.815(\pm0.006)$          &  $0.813(\pm0.007)$          & $0.813(\pm0.005)$ \\  
        &                                                               & CLS+SB & $0.813(\pm0.004)$  &  $0.812(\pm0.007)$          &  $0.814(\pm0.001)$          & $0.814(\pm0.007)$ \\ \hline
    \end{tabular}
    \end{adjustwidth}
\end{table}

\subsection{A5A ablation scores}

% Scores
\begin{figure}[H]
    \begin{adjustwidth}{-1.5cm}{-1.5cm}
    \centering
        \begin{subfigure}{0.4\textwidth}
            \centering
            \includegraphics[width=\textwidth]{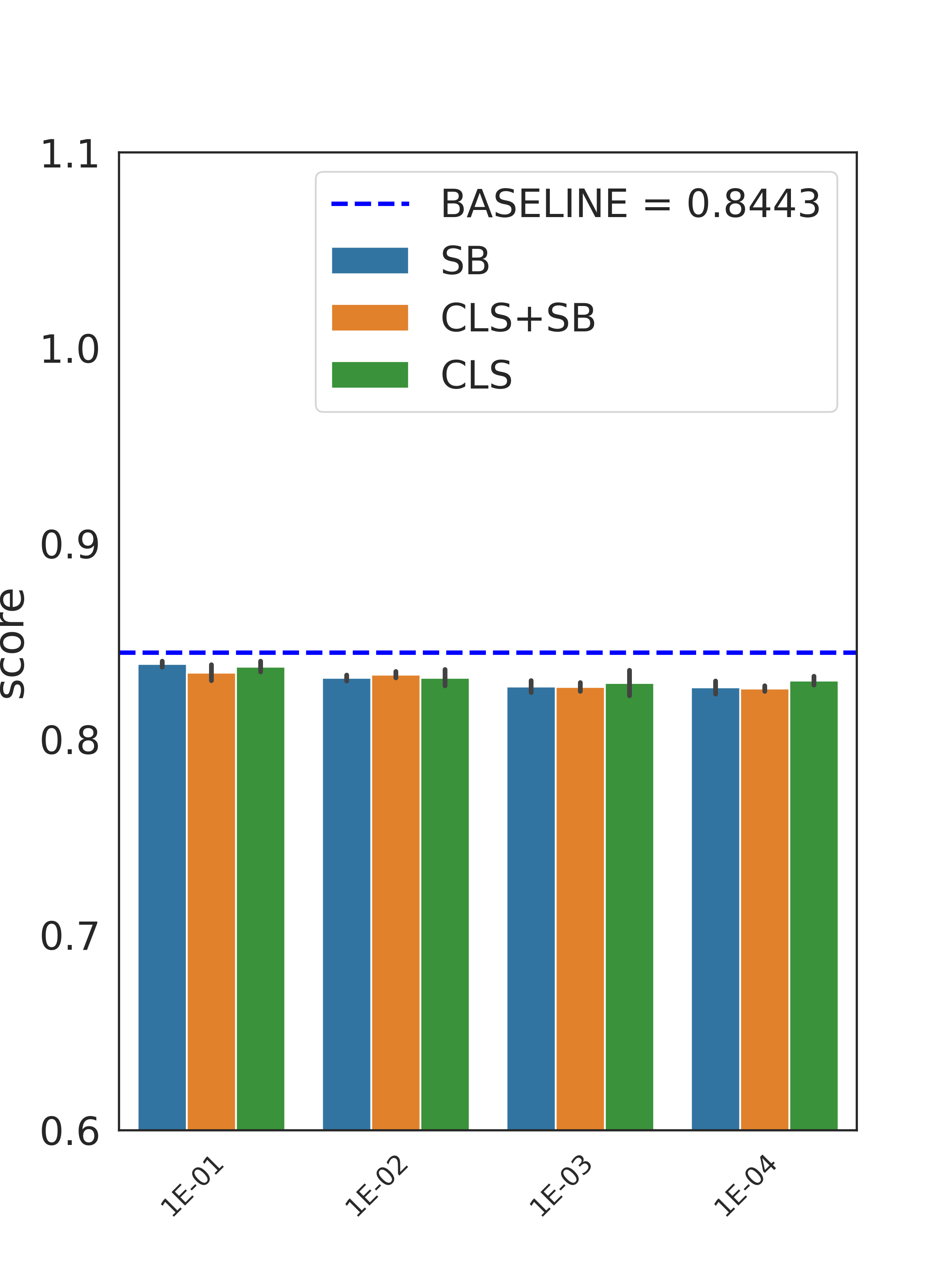}
            \caption{0\% noise}
            \label{fig:a5a_score_0_ablation}
        \end{subfigure}\hfill
        \begin{subfigure}{0.4\textwidth}
            \centering
            \includegraphics[width=\textwidth]{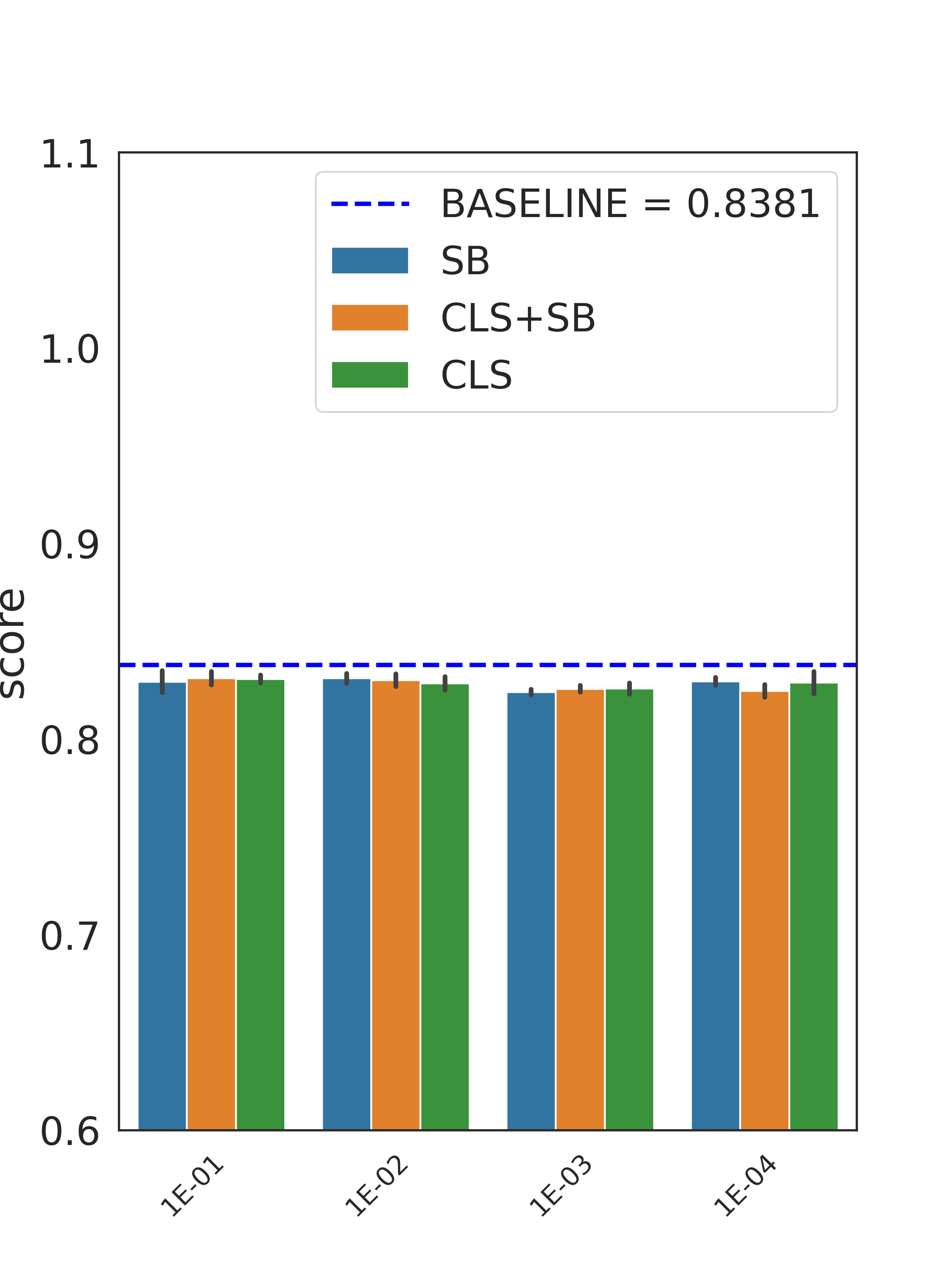}
            \caption{15\% noise}
            \label{fig:a5a_score_15_ablation}
        \end{subfigure}\hfill
        \begin{subfigure}{0.4\textwidth}
            \centering
            \includegraphics[width=\textwidth]{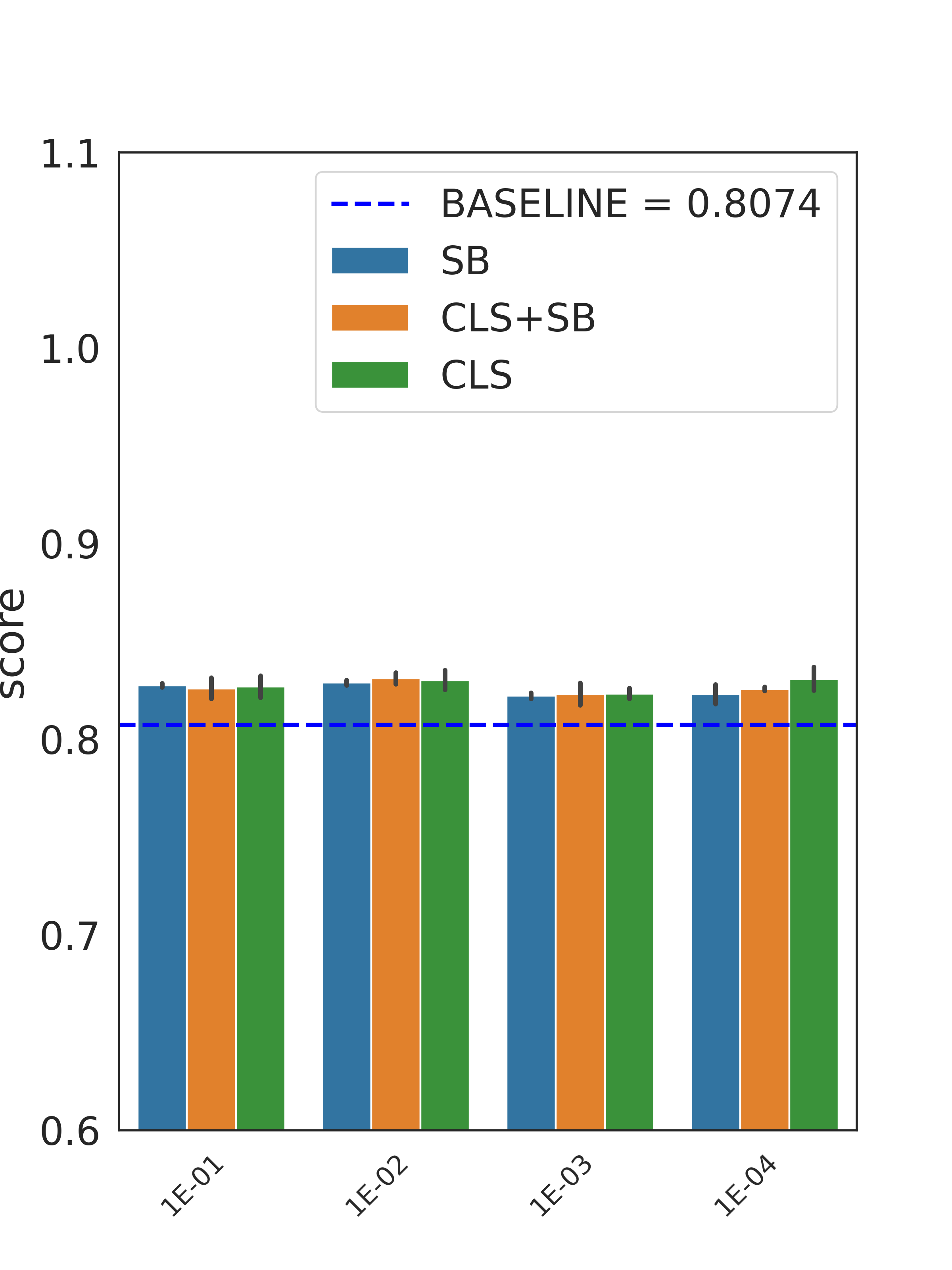}
            \caption{30\% noise}
            \label{fig:a5a_score_3_ablation}
        \end{subfigure}
    \caption{Final Scores for the agents trained over the A5A dataset making the ablations.}
    \label{fig:a5a_scores_ablations}
    \end{adjustwidth}
\end{figure}

\begin{table}[H]
    \centering
    \footnotesize
    \begin{adjustwidth}{-1.5cm}{-1.5cm}
    \begin{tabular}{c|l|l||l|l|l|l}
    \hline
    \textbf{Data} & \multicolumn{1}{c|}{\textbf{Noise}} & \diagbox{\textbf{Method}}{\textbf{C2}}            & \multicolumn{1}{c|}{\textbf{1e-1}}   & \multicolumn{1}{c|}{\textbf{1e-2}}               & \multicolumn{1}{c|}{\textbf{1e-3}}               & \multicolumn{1}{c}{\textbf{1e-4}} \\ \hline
    \multirow{9}{*}{A5A} & {\multirow{3}{*}{0}}                         & SB     & $\mathbf{0.838(\pm 0.002)}$          & $0.831(\pm 0.002)$           & $0.827(\pm 0.003)$                 & $0.826(\pm 0.004)$ \\  
        &                                                               & CLS    & $0.837(\pm 0.003)$                   & $0.831(\pm 0.005)$           & $0.829(\pm 0.007)$                 & $0.830(\pm 0.003)$ \\  
        &                                                               & CLS+SB & $0.834(\pm 0.004)$                   & $0.833(\pm 0.002)$           & $0.827(\pm 0.003)$                 & $0.826(\pm 0.002)$ \\ \cline{2-7} 
        
        & \multirow{3}{*}{15}                                           & SB     & $0.829(\pm 0.006)$                   & $0.831(\pm 0.003)$           &  $0.824(\pm 0.002)$                & $0.830(\pm 0.002)$ \\  
        &                                                               & CLS    & $\mathbf{0.831(\pm 0.002)}$          & $0.828(\pm 0.004)$           &  $0.826(\pm 0.003)$                & $0.829(\pm 0.006)$ \\  
        &                                                               & CLS+SB & $0.831(\pm 0.004)$                   & $0.830(\pm 0.004)$           &  $0.826(\pm 0.002)$                & $0.825(\pm 0.004)$ \\ \cline{2-7} 
        
        & \multirow{3}{*}{30}                                           & SB     & $0.828(\pm 0.001)$                   & $\mathbf{0.829(\pm 0.001)}$  &  $0.822(\pm 0.002)$                & $0.823(\pm 0.006)$ \\  
        &                                                               & CLS    & $0.827(\pm 0.006)$                   & $0.830(\pm 0.006)$           &  $0.823(\pm 0.003)$                & $0.831(\pm 0.007)$ \\  
        &                                                               & CLS+SB & $0.826(\pm 0.006)$                   & $\mathbf{0.831(\pm 0.003)}$  &  $0.823(\pm 0.006)$                & $0.826(\pm 0.001)$ \\ \hline
    \end{tabular}
    \end{adjustwidth}
\end{table}

\subsection{W5A ablation scores}
% Scores
\begin{figure}[H]
    \begin{adjustwidth}{-1.5cm}{-1.5cm}
    \centering
        \begin{subfigure}{0.4\textwidth}
            \centering
            \includegraphics[width=\textwidth]{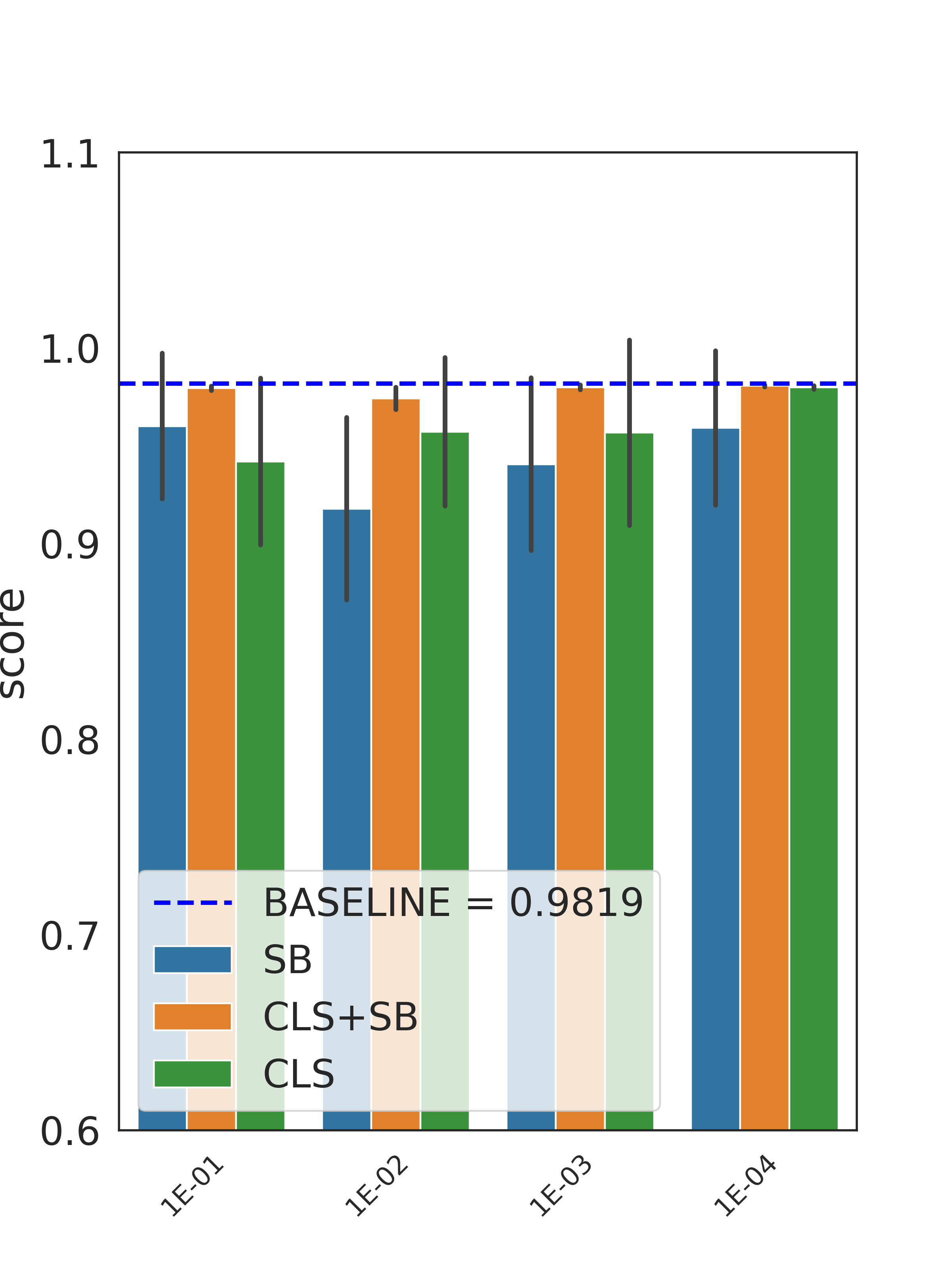}
            \caption{0\% noise}
            \label{fig:w5a_score_0_ablation}
        \end{subfigure}\hfill
        \begin{subfigure}{0.4\textwidth}
            \centering
            \includegraphics[width=\textwidth]{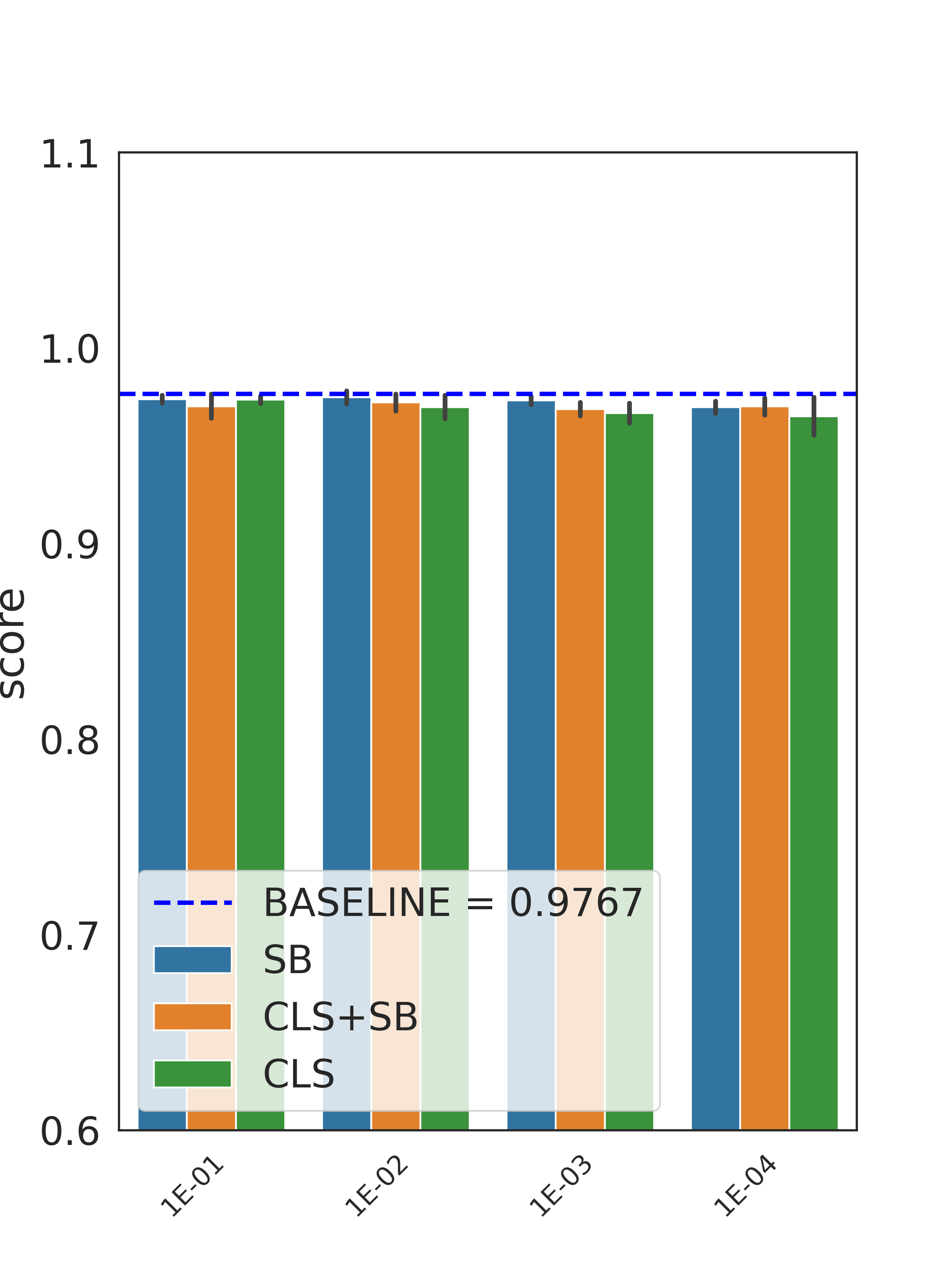}
            \caption{15\% noise}
            \label{fig:w5a_score_15_ablation}
        \end{subfigure}\hfill
        \begin{subfigure}{0.4\textwidth}
            \centering
            \includegraphics[width=\textwidth]{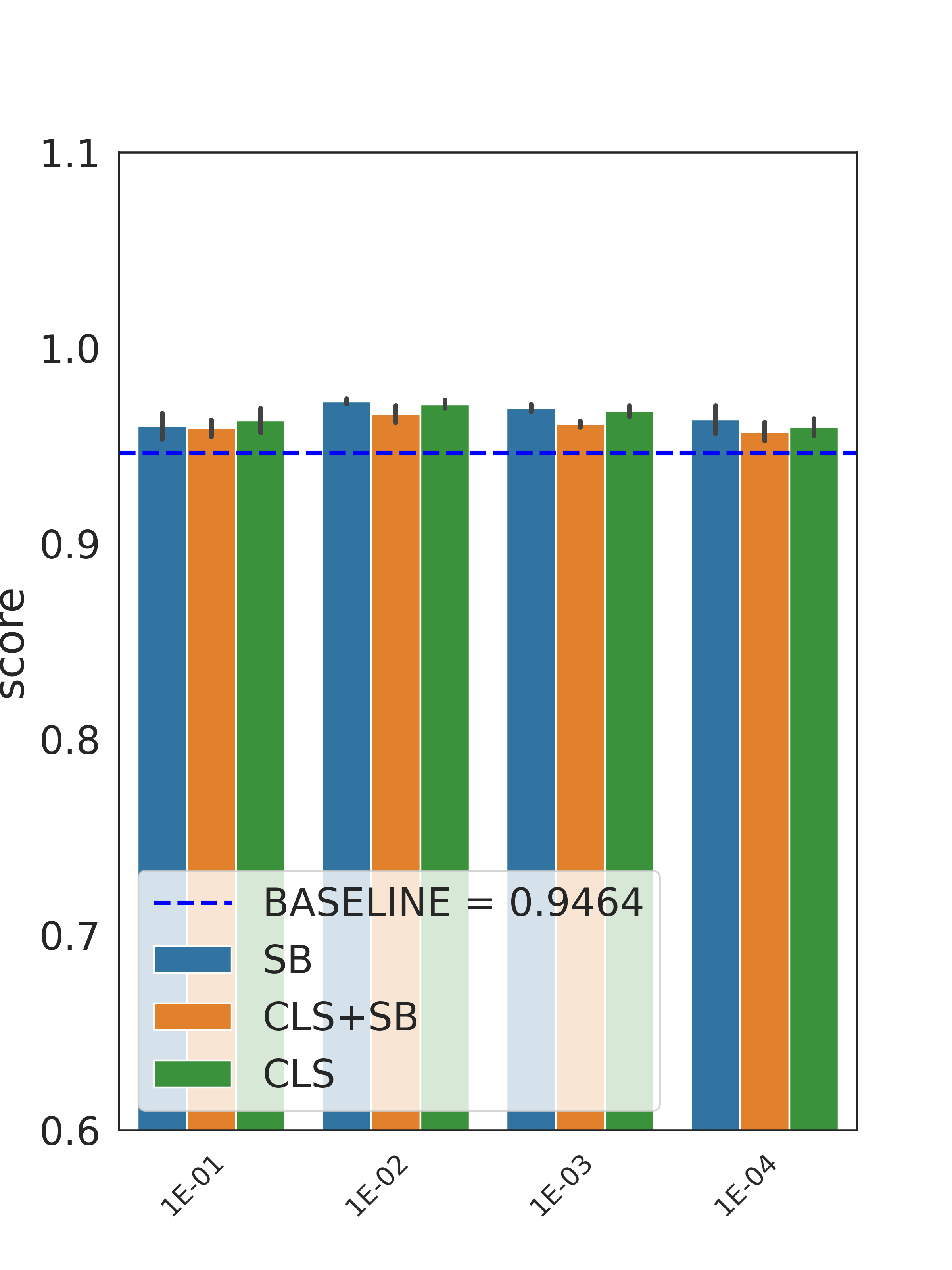}
            \caption{30\% noise}
            \label{fig:w5a_score_3_ablation}
        \end{subfigure}
    \caption{Final Scores for the agents trained over the W5A dataset making the ablations.}
    \label{fig:w5a_scores_ablations}
    \end{adjustwidth}
\end{figure}

\begin{table}[H]
    \centering
    \footnotesize
    \begin{adjustwidth}{-1.5cm}{-1.5cm}
    \begin{tabular}{c|l|l||l|l|l|l}
    \hline
    \textbf{Data} & \multicolumn{1}{c|}{\textbf{Noise}} & \diagbox{\textbf{Method}}{\textbf{C2}}            & \multicolumn{1}{c|}{\textbf{1e-1}}   & \multicolumn{1}{c|}{\textbf{1e-2}}               & \multicolumn{1}{c|}{\textbf{1e-3}}               & \multicolumn{1}{c}{\textbf{1e-4}} \\ \hline
    \multirow{9}{*}{W5A} & {\multirow{3}{*}{0}}                         & SB     & $0.960(\pm 0.042)$                   & $0.918(\pm 0.052)$           & $0.941(\pm 0.049)$                 & $0.959(\pm 0.044)$ \\  
        &                                                               & CLS    & $0.942(\pm 0.048)$                   & $0.957(\pm 0.042)$           & $0.957(\pm 0.053)$                 & $0.980(\pm 0.001)$ \\  
        &                                                               & CLS+SB & $0.979(\pm 0.001)$                   & $0.974(\pm 0.006)$           & $0.980(\pm 0.001)$                 & $\mathbf{0.981(\pm 0.001)}$ \\ \cline{2-7} 
        
        & \multirow{3}{*}{15}                                           & SB     & $\mathbf{0.974(\pm 0.002)}$          & $0.975(\pm 0.004)$           & $0.973(\pm 0.002)$                 & $0.970(\pm 0.004)$ \\  
        &                                                               & CLS    & $0.973(\pm 0.002)$                   & $0.970(\pm 0.007)$           & $0.967(\pm 0.006)$                 & $0.965(\pm 0.011)$ \\  
        &                                                               & CLS+SB & $0.970(\pm 0.007)$                   & $0.972(\pm 0.005)$           & $0.969(\pm 0.004)$                 & $0.970(\pm 0.005)$ \\ \cline{2-7} 
        
        & \multirow{3}{*}{30}                                           & SB     & $0.960(\pm 0.008)$                   & $\mathbf{0.973(\pm 0.001)}$  & $0.969(\pm 0.002)$                 & $0.963(\pm 0.008)$ \\  
        &                                                               & CLS    & $0.963(\pm 0.007)$                   & $0.971(\pm 0.002)$           & $0.968(\pm 0.003)$                 & $0.960(\pm 0.005)$ \\  
        &                                                               & CLS+SB & $0.959(\pm 0.005)$                   & $0.966(\pm 0.005)$           & $0.961(\pm 0.002)$                 & $0.957(\pm 0.005)$ \\ \hline
    \end{tabular}
    \end{adjustwidth}
\end{table}

\subsection{CIFAR10 ablation scores}
% Scores
\begin{figure}[H]
    \begin{adjustwidth}{-1.5cm}{-1.5cm}
    \centering
        \begin{subfigure}{0.4\textwidth}
            \centering
            \includegraphics[width=\textwidth]{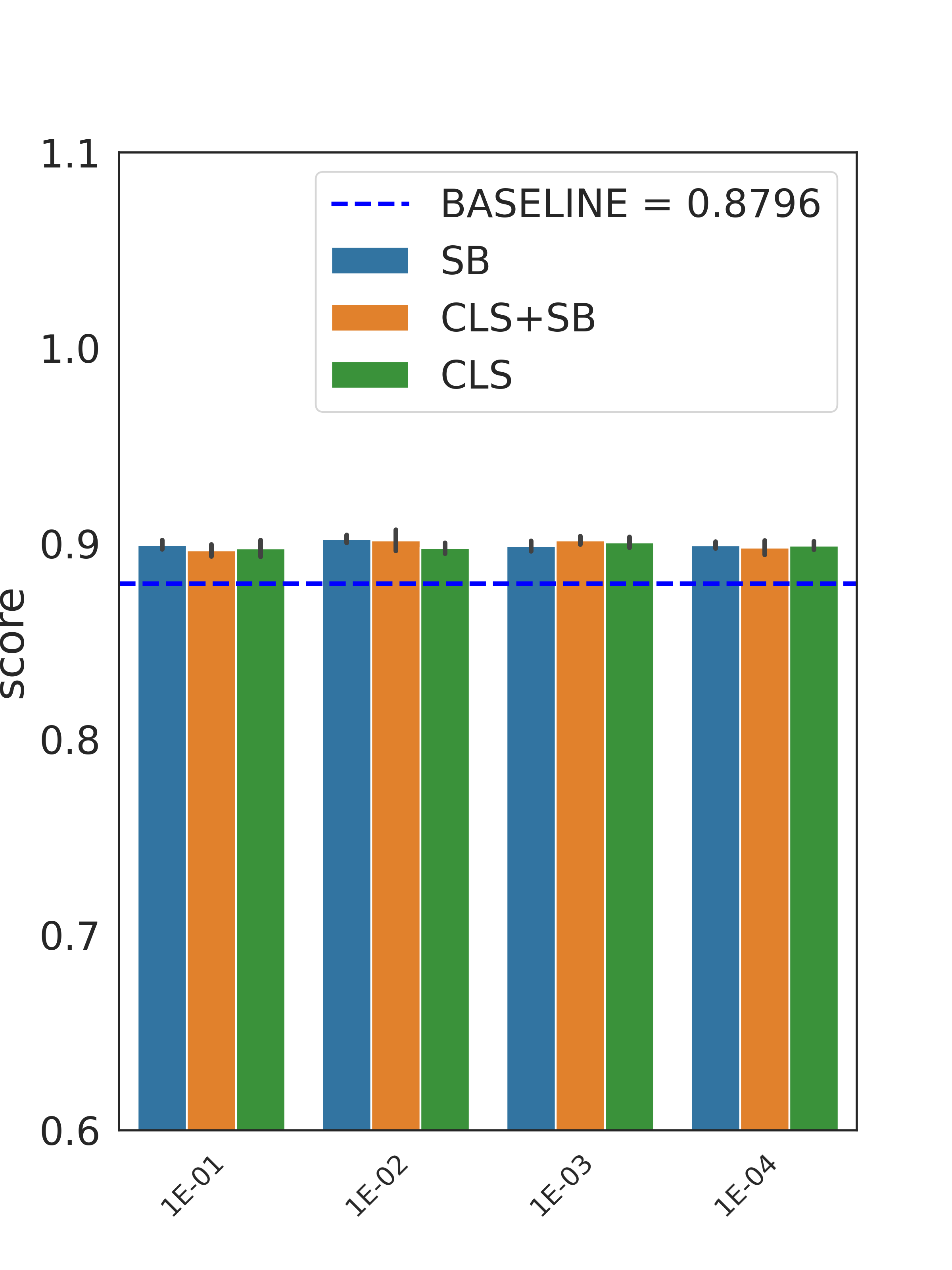}
            \caption{0\% noise}
            \label{fig:cifat10_score_0_ablation}
        \end{subfigure}\hfill
        \begin{subfigure}{0.4\textwidth}
            \centering
            \includegraphics[width=\textwidth]{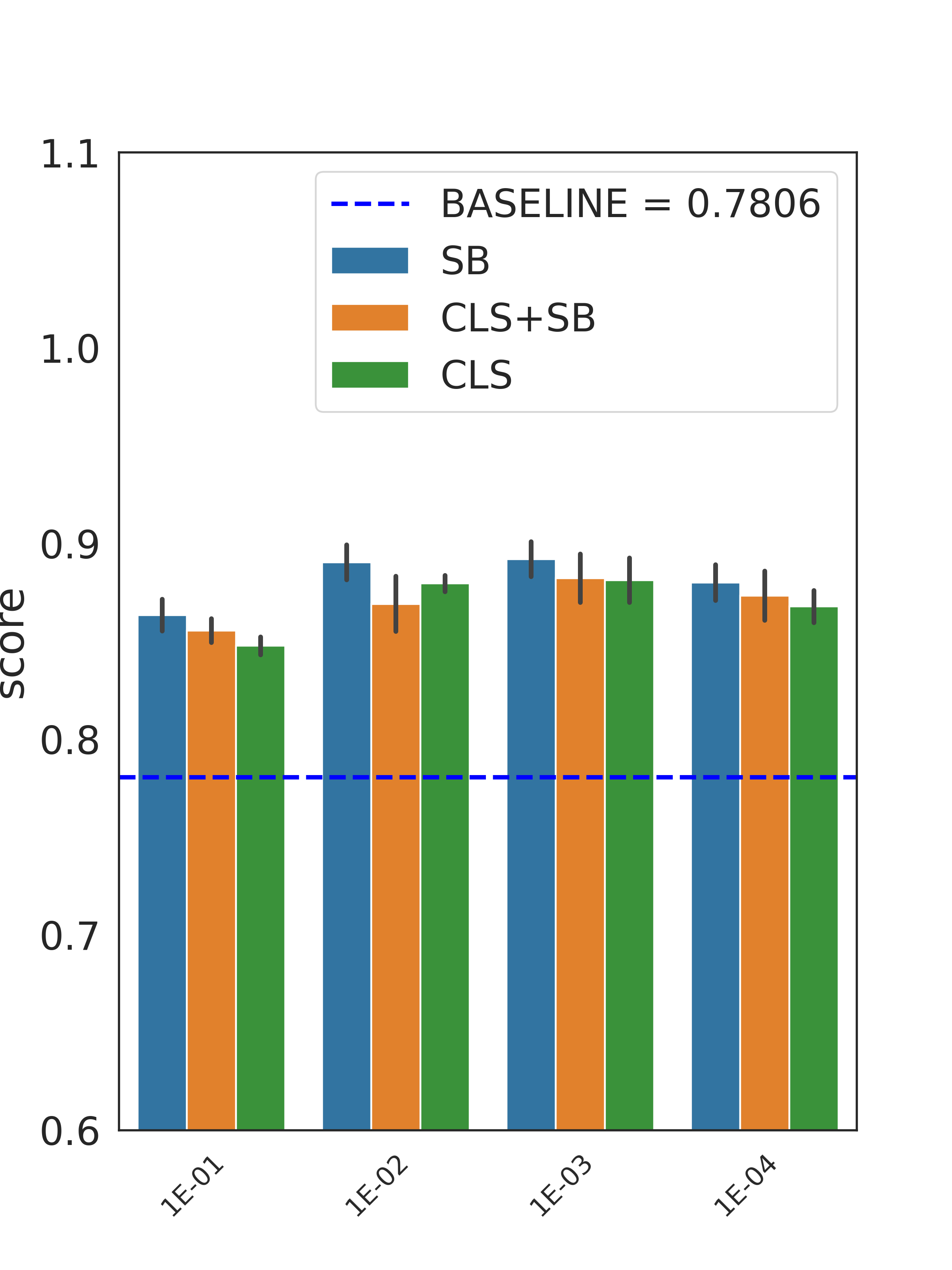}
            \caption{15\% noise}
            \label{fig:cifar10_score_15_ablation}
        \end{subfigure}\hfill
        \begin{subfigure}{0.4\textwidth}
            \centering
            \includegraphics[width=\textwidth]{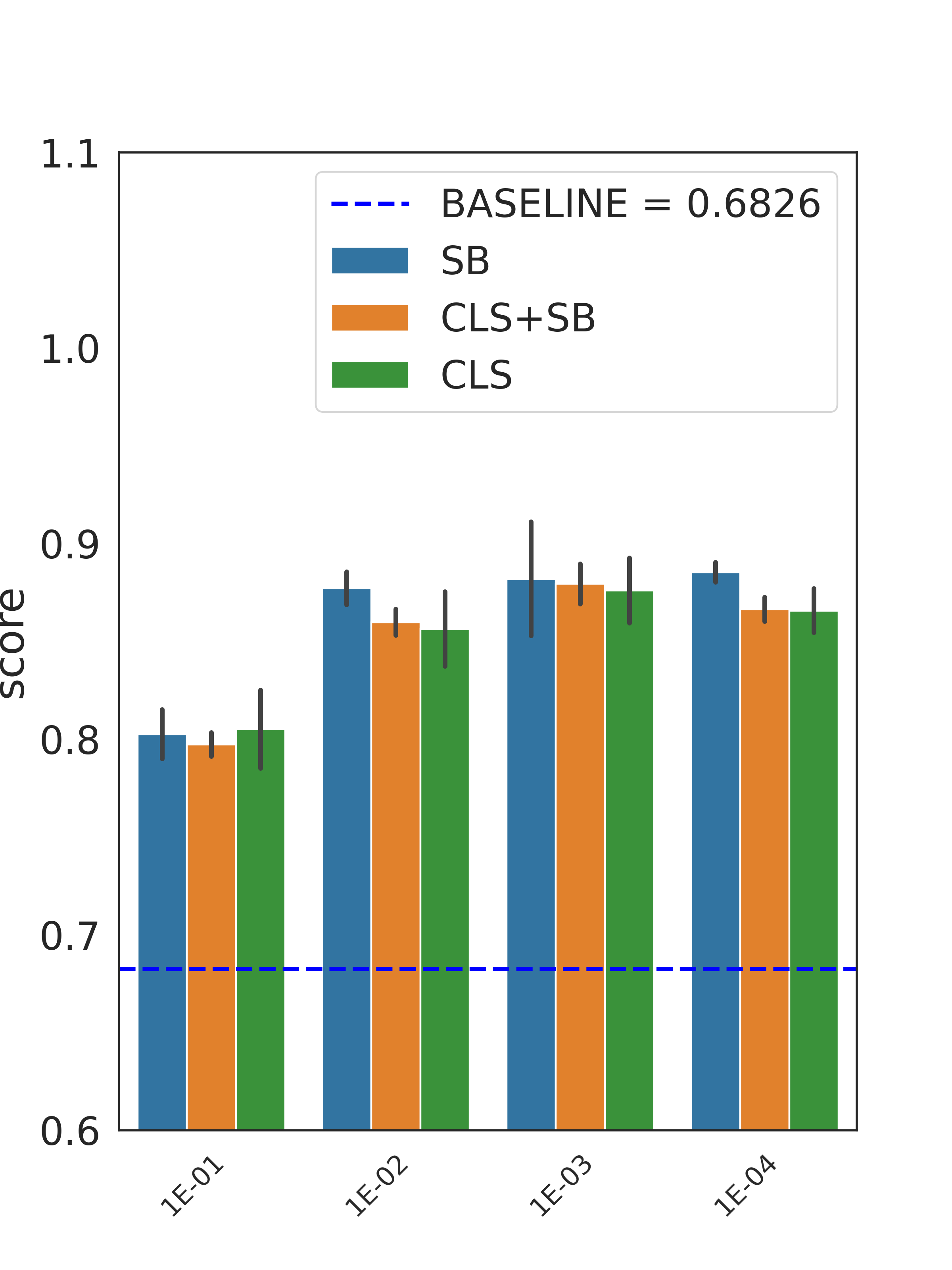}
            \caption{30\% noise}
            \label{fig:cifar10_score_3_ablation}
        \end{subfigure}
    \caption{Final Scores for the agents trained over the CIFAR10 dataset making the ablations}
    \label{fig:cifar10_scores_ablations}
    \end{adjustwidth}
\end{figure}

\begin{table}[H]
    \centering
    \footnotesize
    \begin{adjustwidth}{-1.5cm}{-1.5cm}
    \begin{tabular}{c|l|l||l|l|l|l}
    \hline
    \textbf{Data} & \multicolumn{1}{c|}{\textbf{Noise}} & \diagbox{\textbf{Method}}{\textbf{C2}}            & \multicolumn{1}{c|}{\textbf{1e-1}}      & \multicolumn{1}{c|}{\textbf{1e-2}}               & \multicolumn{1}{c|}{\textbf{1e-3}}               & \multicolumn{1}{c}{\textbf{1e-4}} \\ \hline
    \multirow{9}{*}{CIFAR10} & {\multirow{3}{*}{0}}                     & SB                                & $0.899(\pm0.003)$                       & $\mathbf{0.902(\pm0.002)}$                       & $0.899(\pm0.003)$                                & $0.899(\pm0.002)$ \\  
        &                                                               & CLS                               & $0.897(\pm0.005)$                       & $0.898(\pm0.003)$                                & $0.901(\pm0.003)$                                & $0.899(\pm0.002)$ \\  
        &                                                               & CLS+SB                            & $0.896(\pm0.003)$                       & $0.902(\pm0.006)$                                & $\mathbf{0.902(\pm0.002)}$                       & $0.898(\pm0.004)$ \\ \cline{2-7} 
        
        & \multirow{3}{*}{15}                                           & SB                                & $0.863(\pm0.009)$                       & $0.89(\pm0.01)$                                  & $\mathbf{0.892(\pm0.01)}$                        & $0.88(\pm0.01)$\\  
        &                                                               & CLS                               & $0.848(\pm0.005)$                       & $0.88(\pm0.005)$                                 & $0.881(\pm0.013)$                                & $0.868(\pm0.009)$ \\  
        &                                                               & CLS+SB                            & $0.855(\pm0.007)$                       & $0.869(\pm0.016)$                                & $0.882(\pm0.014)$                                & $0.873(\pm0.014)$  \\ \cline{2-7} 
        
        & \multirow{3}{*}{30}                                           & SB                                & $0.805(\pm0.022)$                       & $0.856(\pm0.021)$                                & $0.876(\pm0.019)$                                & $0.866(\pm0.013)$   \\  
        &                                                               & CLS                               & $0.797(\pm0.007)$                       & $0.86(\pm0.007)$                                 & $0.879(\pm0.011)$                                & $0.866(\pm0.007)$  \\  
        &                                                               & CLS+SB                            & $0.802(\pm0.014)$                       & $0.877(\pm0.009)$                                & $0.882(\pm0.033)$                                & $\mathbf{0.885(\pm0.006)}$  \\ \hline
    \end{tabular}
    \end{adjustwidth}
\end{table}

\subsection{MNIST ablation scores}
% Scores
\begin{figure}[H]
    \begin{adjustwidth}{-1.5cm}{-1.5cm}
    \centering
        \begin{subfigure}{0.4\textwidth}
            \centering
            \includegraphics[width=\textwidth]{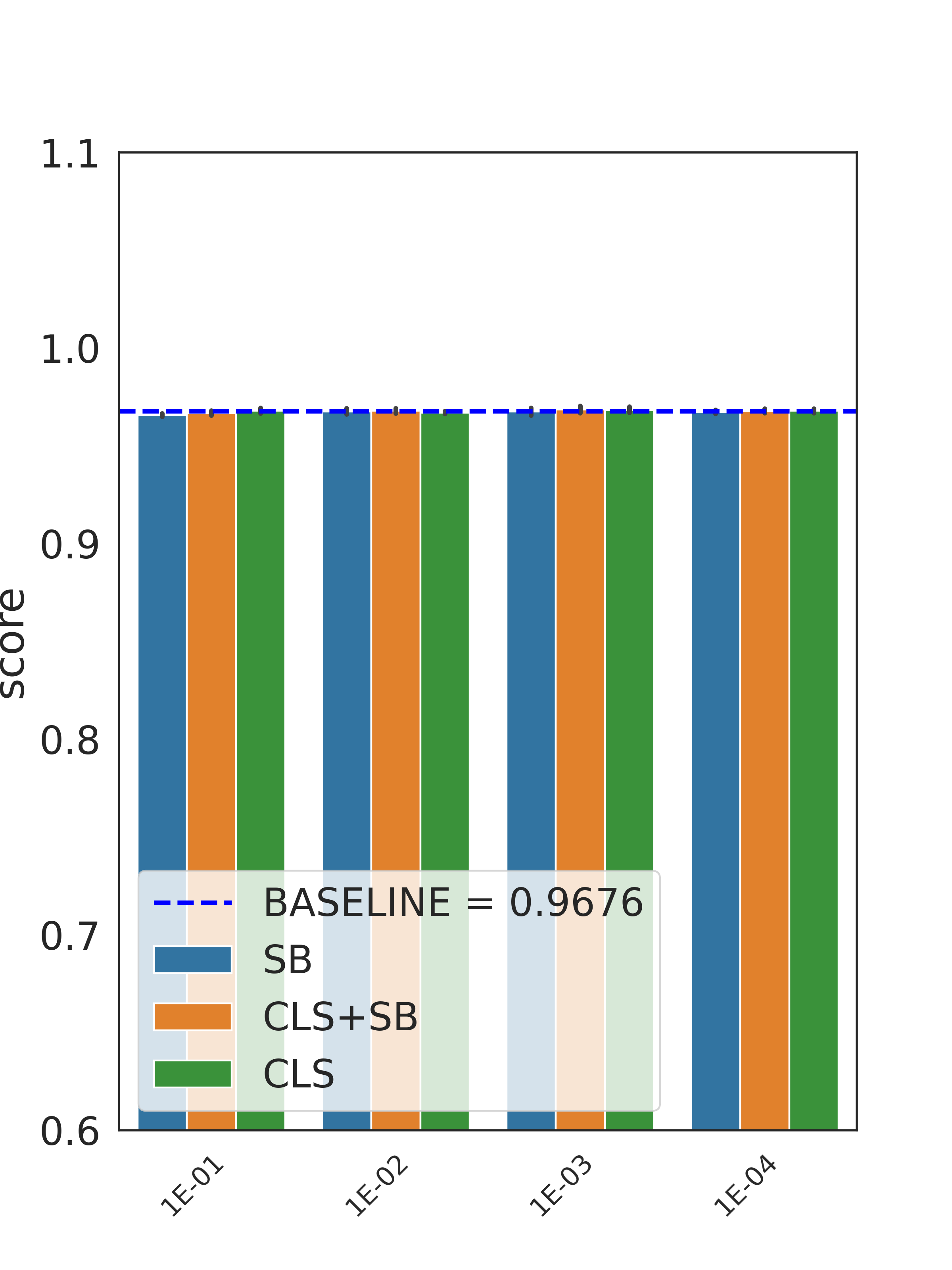}
            \caption{0\% noise}
            \label{fig:mnist_score_0_ablation}
        \end{subfigure}\hfill
        \begin{subfigure}{0.4\textwidth}
            \centering
            \includegraphics[width=\textwidth]{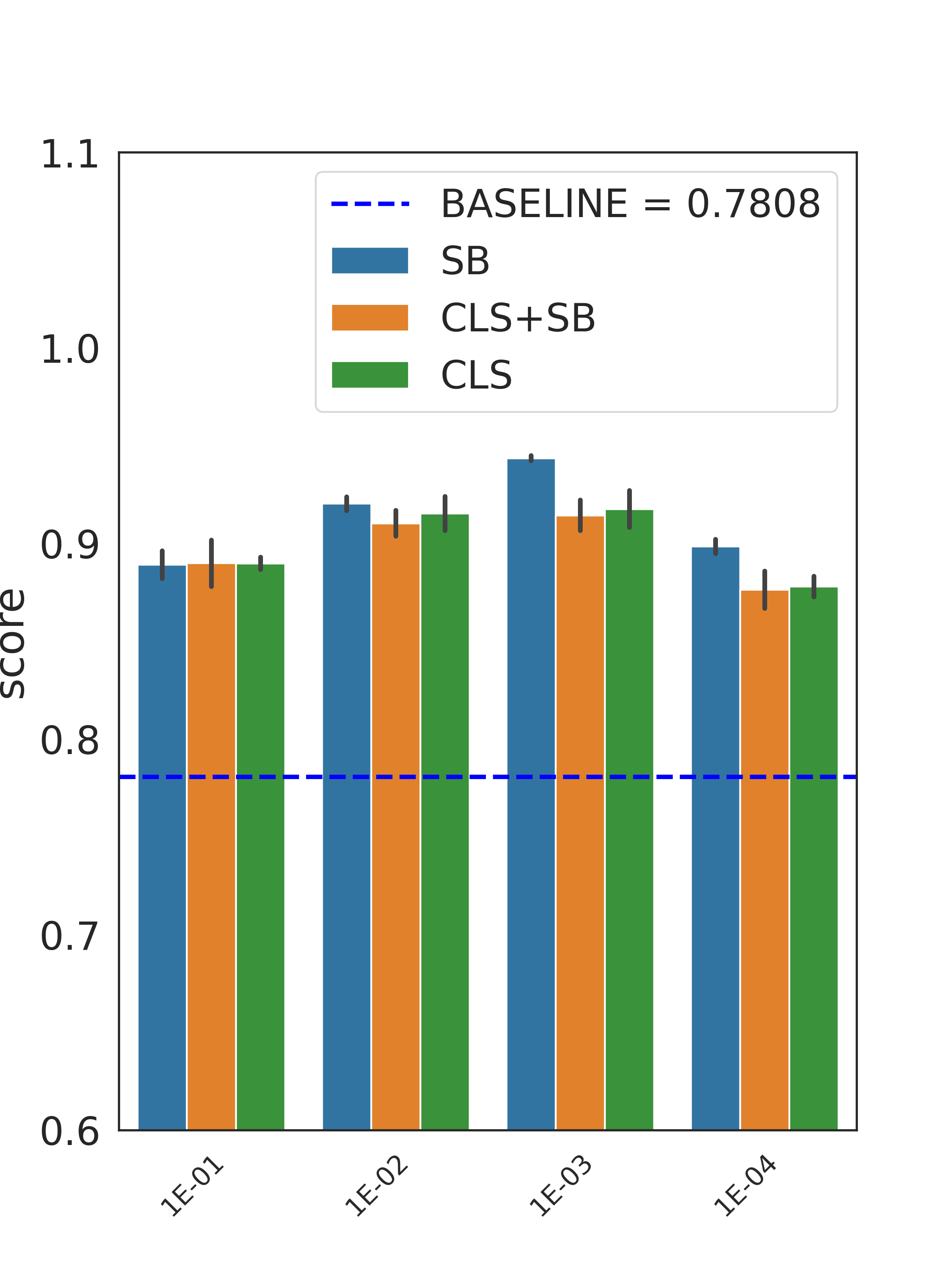}
            \caption{15\% noise}
            \label{fig:mnist_score_15_ablation}
        \end{subfigure}\hfill
        \begin{subfigure}{0.4\textwidth}
            \centering
            \includegraphics[width=\textwidth]{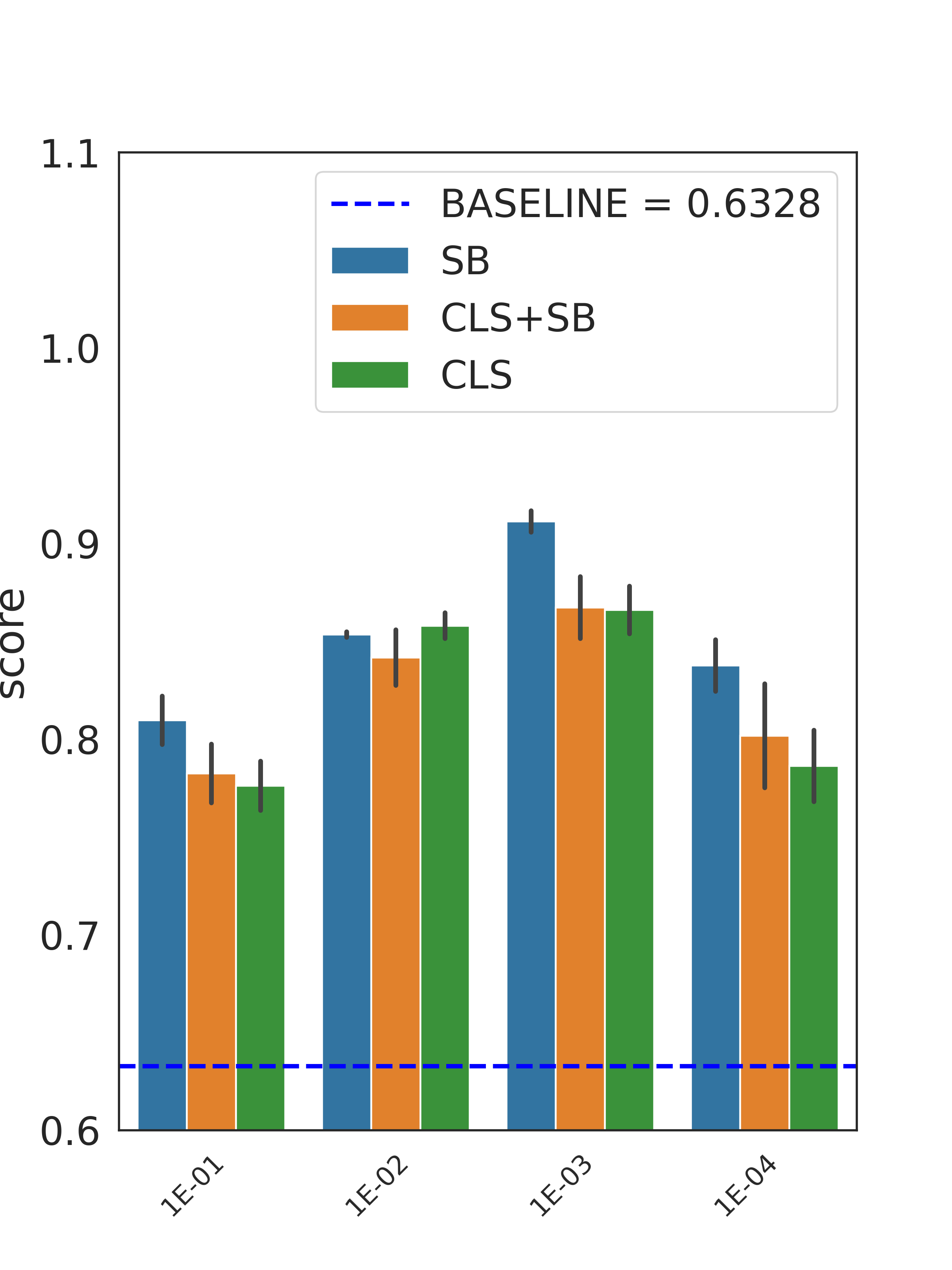}
            \caption{30\% noise}
            \label{fig:mnist_score_3_ablation}
        \end{subfigure}
    \caption{Final Scores for the agents trained over the MNIST dataset making the ablations}
    \label{fig:mnist_scores_ablations}
    \end{adjustwidth}
\end{figure}

\begin{table}[H]
    \centering
    \footnotesize
    \begin{adjustwidth}{-1.5cm}{-1.5cm}
    \begin{tabular}{c|l|l||l|l|l|l}
    \hline
    \textbf{Data} & \multicolumn{1}{c|}{\textbf{Noise}} & \diagbox{\textbf{Method}}{\textbf{C2}}            & \multicolumn{1}{c|}{\textbf{1e-1}}      & \multicolumn{1}{c|}{\textbf{1e-2}}               & \multicolumn{1}{c|}{\textbf{1e-3}}               & \multicolumn{1}{c}{\textbf{1e-4}} \\ \hline
    \multirow{9}{*}{MNIST} & {\multirow{3}{*}{0}}                       & SB                                &  $0.966(\pm0.001)$                      &  $0.967(\pm0.002)$                               & $0.967(\pm0.002)$                                & $\mathbf{0.968(\pm0.001)}$ \\  
        &                                                               & CLS                               &  $0.967(\pm0.002)$                      &  $0.967(\pm0.001)$                               & $\mathbf{0.968(\pm0.001)}$                       & $\mathbf{0.968(\pm0.001)}$ \\  
        &                                                               & CLS+SB                            &  $0.967(\pm0.001)$                      &  $0.968(\pm0.002)$                               & $\mathbf{0.969(\pm0.002)}$                       & $\mathbf{0.968(\pm0.001)}$ \\ \cline{2-7} 
        
        & \multirow{3}{*}{15}                                           & SB                                & $0.881(\pm0.013)$                       &  $0.909(\pm0.017)$                               & $\mathbf{0.941(\pm0.004)}$                       & $0.891(\pm0.011)$ \\  
        &                                                               & CLS                               & $0.893(\pm0.005)$                       &  $0.915(\pm0.008)$                               & $0.919(\pm0.009)$                                & $0.875(\pm0.01)$ \\  
        &                                                               & CLS+SB                            & $0.89(\pm0.013)$                        &  $0.91(\pm0.007)$                                & $0.914(\pm0.009)$                                & $0.876(\pm0.011)$ \\ \cline{2-7} 
        
        & \multirow{3}{*}{30}                                           & SB                                & $0.803(\pm0.014)$                       &  $0.848(\pm0.011)$                               & $\mathbf{0.913(\pm0.01)}$                        & $0.834(\pm0.016)$ \\  
        &                                                               & CLS                               & $0.775(\pm0.023)$                       &  $0.85(\pm0.014)$                                & $0.865(\pm0.019)$                                & $0.797(\pm0.023)$ \\  
        &                                                               & CLS+SB                            & $0.782(\pm0.017)$                       &  $0.842(\pm0.016)$                               & $0.867(\pm0.018)$                                & $0.802(\pm0.03)$ \\ \hline
    \end{tabular}
    \end{adjustwidth}
\end{table}

\section{Time cost comparison}\label{appendix:time}
\begin{figure}[H]
    \begin{adjustwidth}{-2cm}{-2cm}
        \centering
        \begin{subfigure}{0.65\textwidth}
            \centering
            \includegraphics[width=\textwidth]{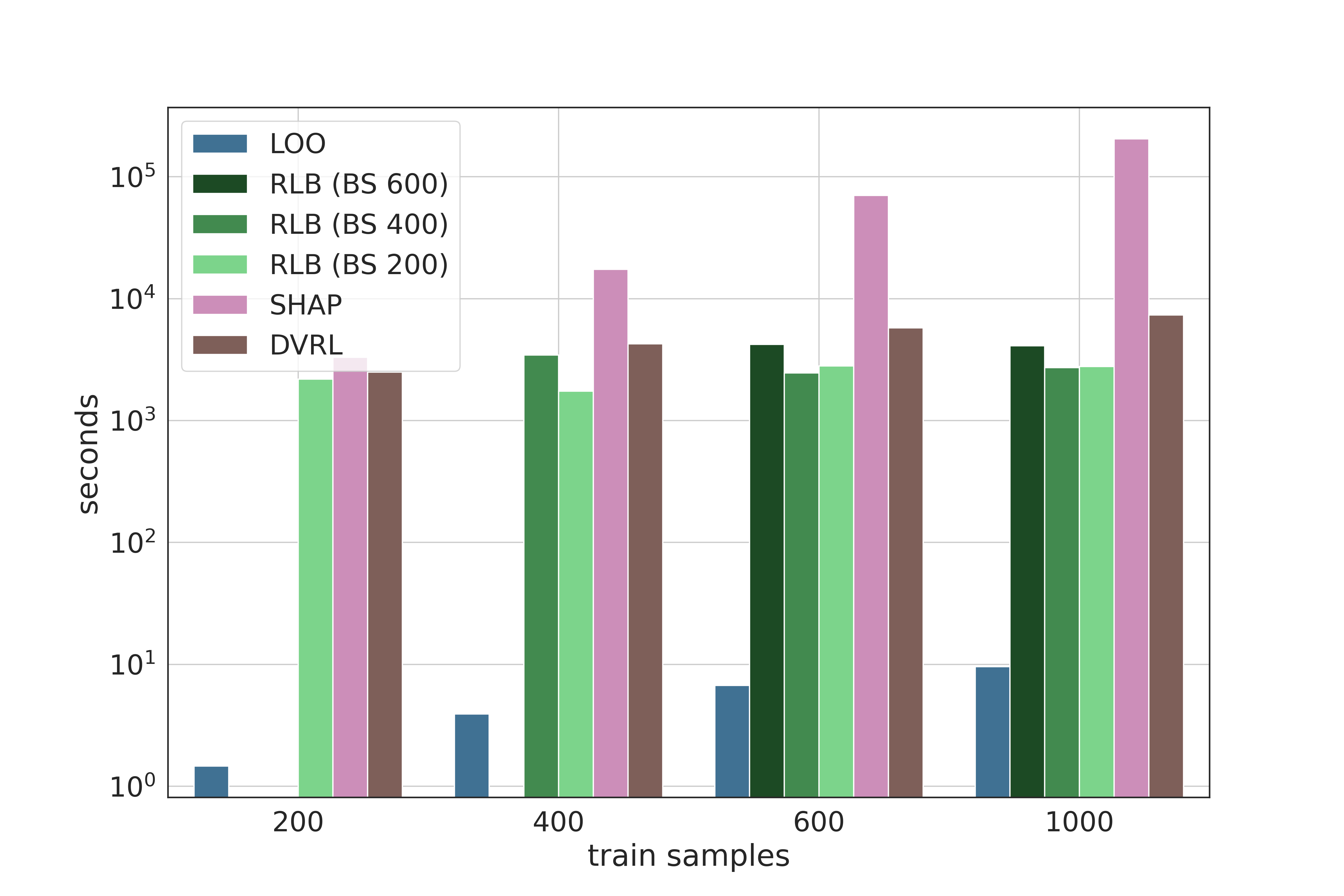}
            \caption{Time comparison with LOO data.}
            \label{fig:time_with_loo}
        \end{subfigure}
        \hfill
        \begin{subfigure}{0.65\textwidth}
            \centering
            \includegraphics[width=\textwidth]{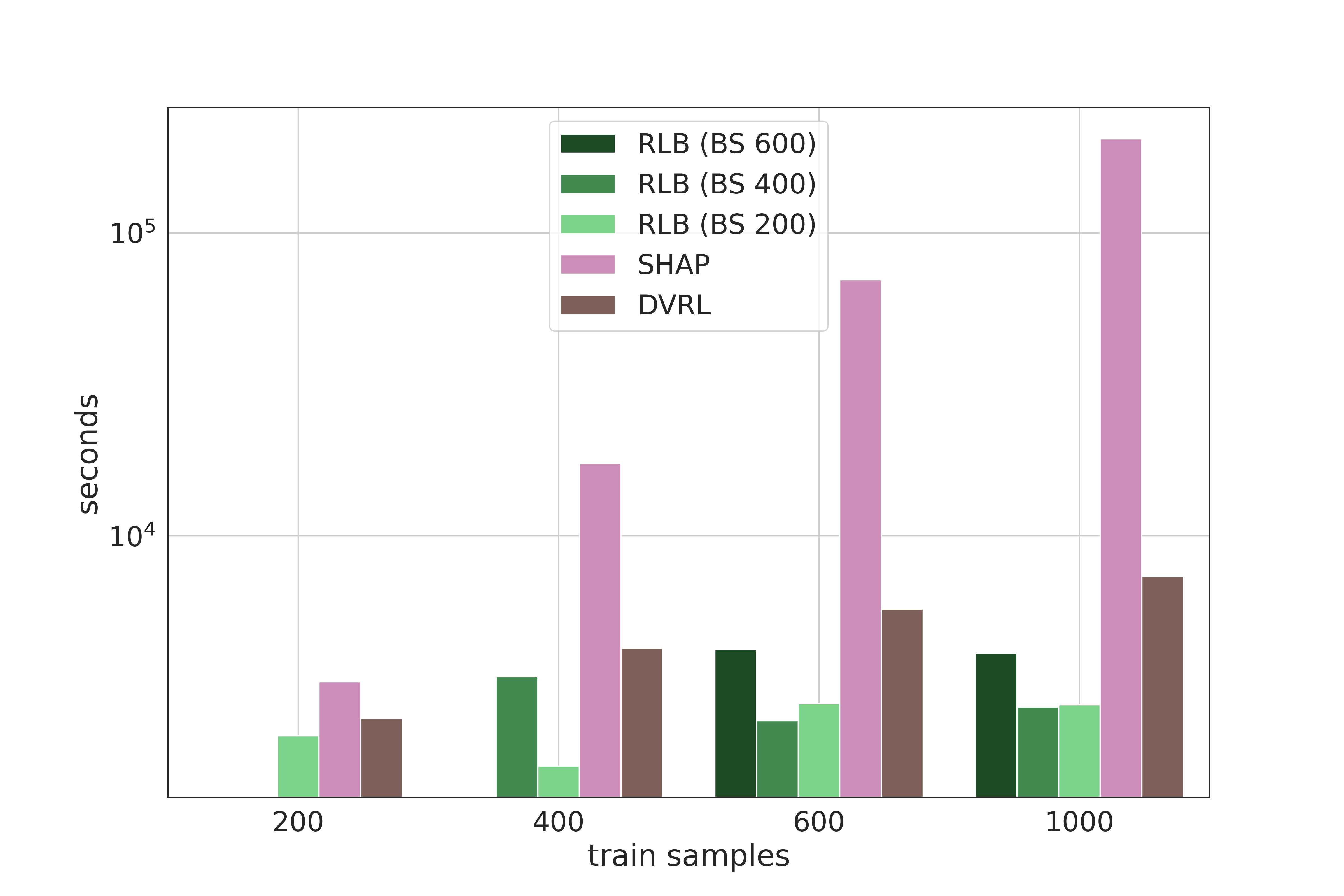}
            \caption{Time comparison without LOO data.}
            \label{fig:time_without_loo}
        \end{subfigure} 
    \caption{Time comparisons between methods}
    \label{fig:time_comparison}
    \end{adjustwidth}
\end{figure}

As this is a reinforcement learning problem, the computational cost is increasing in the number of interactions of the environment with the agent. The purpose of the Figure \ref{fig:time_comparison} is to illustrate several concepts. First, the high computational cost of Data Shapley and the low computational cost of LOO. Secondly, the stability of the computational cost of our agents regardless of the number of training samples we are using. And thirdly and lastly the cost of increasing the batch size of examples that see each interaction of the agent with the environment.

\end{appendices}

\bibliography{ms}{}
\bibliographystyle{plainnat}

\end{document}